\documentclass[10pt,twocolumn,letterpaper]{article}

\usepackage[pagenumbers]{wacv} 

\definecolor{wacvblue}{rgb}{0.21,0.49,0.74}
\usepackage[pagebackref,breaklinks,colorlinks,allcolors=wacvblue]{hyperref}

\def\wacvPaperID{2401} 
\def\confName{WACV}
\def\confYear{2026}

\usepackage[utf8]{inputenc} 
\usepackage[T1]{fontenc}    
\usepackage{hyperref}       
\usepackage{url}            
\usepackage{booktabs}       
\usepackage{amsfonts}       
\usepackage{nicefrac}       
\usepackage{microtype}      
\usepackage{xcolor}         

\usepackage{color}
\usepackage{colortbl}
\usepackage{amsmath}
\usepackage{algorithm}
\usepackage{algorithmic}
\usepackage{multirow}

\usepackage[capitalize,noabbrev]{cleveref}

\usepackage{microtype}
\usepackage{graphicx}
\usepackage{subcaption}
\usepackage{caption}

\title{DPBridge: Latent Diffusion Bridge for Dense Prediction}

\author{Haorui Ji \quad Taojun Lin \quad Hongdong Li\\
The Australian National University\\
{\tt\small \{haorui.ji, taojun.lin, hongdong.li\}@anu.edu.au}
}

\begin{document}

\twocolumn[{%
\renewcommand\twocolumn[1][]{#1}%
\maketitle
\vspace{-3em}
\begin{center}
    \centering
    \captionsetup{type=figure}
    \includegraphics[width=\textwidth]{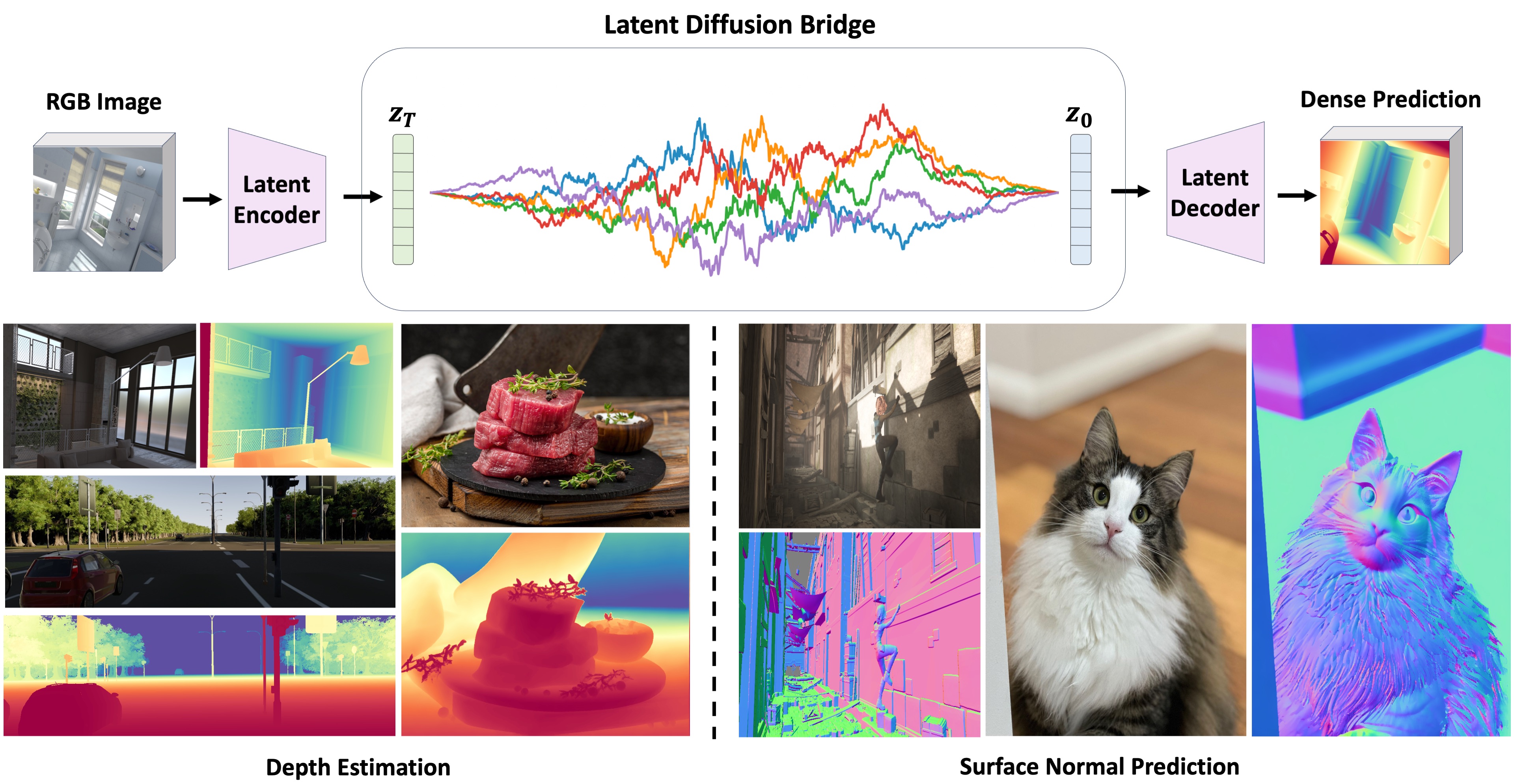}
    \captionof{figure}{We present \textbf{DPBridge} to tackle general dense prediction tasks by establishing conditional probability transport between RGB images and corresponding dense maps in the latent space based on a tractable diffusion bridge process. }
    \label{fig:teaser}
\end{center}%
}]

\vspace{-2em}
\begin{abstract}
Diffusion models demonstrate remarkable capabilities in capturing complex data distributions and have achieved compelling results in many generative tasks. While they have recently been extended to dense prediction tasks such as depth estimation and surface normal prediction, their full potential in this area remains underexplored. As target signal maps and input images are pixel-wise aligned, the conventional noise-to-data generation paradigm is inefficient, and input images can serve as a more informative prior compared to pure noise. Diffusion bridge models, which support data-to-data generation between two general data distributions, offer a promising alternative, but they typically fail to exploit the rich visual priors embedded in large pretrained foundation models. To address these limitations, we integrate diffusion bridge formulation with structured visual priors and introduce DPBridge, the first latent diffusion bridge framework for dense prediction tasks. To resolve the incompatibility between diffusion bridge models and pretrained diffusion backbones, we propose (1) a tractable reverse transition kernel for the diffusion bridge process, enabling maximum likelihood training scheme; (2) finetuning strategies including distribution-aligned normalization and image consistency loss. Experiments across extensive benchmarks validate that our method consistently achieves superior performance, demonstrating its effectiveness and generalization capability under different scenarios.

\end{abstract}
\vspace{-1.5em}
\section{Introduction}
\label{sec:intro}
Dense prediction refers to a class of problems in which the goal is to assign a prediction value for every pixel in the input image. It encompasses a variety of fundamental tasks, including depth estimation~\cite{yang2024depth,ranftl2021vision,eftekhar2021omnidata}, surface normal prediction~\cite{bae2024rethinking,hu2024metric3d,bae2021estimating}, semantic segmentation~\cite{weber2021step,cheng2022masked,wang2023images}, optical flow~\cite{teed2020raft,huang2022flowformer,weinzaepfel2023croco}, etc.

Dense prediction has traditionally relied on discriminative models~\cite{ranftl2021vision,zheng2021rethinking,bae2024rethinking}, which map RGB inputs to dense signal maps directly via feed-forward manners. While these methods offer high inference speed, they usually struggle in ambiguous scenarios-such as occluded regions or dynamic surface normals of glossy objects. This limitation is largely due to the lack of uncertainty modeling. Recently, generative models such as diffusions have been applied to mitigate these drawbacks by framing dense prediction as a conditional generation task. Leveraging the ability of modeling complex data distributions, diffusions serve as robust priors and deliver promising performance as different tasks~\cite{ke2024repurposing,garcia2024fine,fu2025geowizard}. Nevertheless, diffusion-based generative approaches suffer from inherent limitations rooted in their core mechanism: their denoising process initiates from Gaussian noise—an initial prior that contains minimal information about the target dense signal distribution. In such cases, the distributional gap between initial Gaussian prior and target distribution remains large, necessitating more sampling steps and often reliance on manually designed post-hoc guidance~\cite{chung2022diffusion, song2023pseudoinverse,chung2022improving} or customized generation pipeline~\cite{meng2021sdedit}. This not only increases computational overhead but can also degrade performance and prolong convergence.

Instead of starting from noise, a more effective alternative is to directly initiate the generative process from the input. Since RGB images are pixel-wise aligned with their corresponding dense signal maps, they shall act as highly informative priors. This motivates the exploration of diffusion bridge models~\cite{liu2022let,pavon2021data,de2021diffusion,liu2023learning}, another line of computational frameworks based on stochastic processes with fixed start and end states, designed to model probability flows between two given distributions. We argue that diffusion bridge models are more suitable for dense prediction tasks, as their data-to-data generation paradigm naturally leverages the structural correspondence between inputs and outputs, thereby mitigating inefficiencies inherent in the conventional noise-to-data scheme. Despite theoretical feasibility, diffusion bridge-based methods~\cite{zhou2023denoising,li2023bbdm,liu20232} have not yet demonstrated competitive performance compared with standard diffusion counterparts. This performance gap primarily arises because most bridge models are trained from scratch via score matching objectives, making them incompatible with the widely adopted pretrained diffusion backbones. Consequently, these methods fail to leverage the embedded semantic and spatial prior knowledge, constraining their expressive power and generalization capability in dense prediction scenarios.

In this paper, we propose \textbf{D}ense \textbf{P}rediction \textbf{Bridge} (\textbf{DPBridge}), the first latent diffusion bridge model tailored for general dense prediction tasks by integrating the strengths of diffusion bridge modeling with visual priors from pretrained backbones. Specifically, we derive a fully tractable bridge process to connect the latent representations of RGB images and their corresponding dense signal maps, and formulate the generation process as a sequence of transformations that progressively synthesize the target maps from input data. Furthermore, we introduce finetuning strategies to better adapt pretrained diffusion backbones for DPBridge construction. Leveraging rich visual priors not only enhances perceptual performance and generalization ability, but also significantly accelerates training convergence. \cref{fig:teaser} provides a high-level overview of the DPBridge framework and illustrates its effectiveness across two representative tasks: depth estimation and surface normal prediction. These results demonstrate that our method is both theoretically feasible and practically adaptable to diverse scenarios.

Contributions of this paper can be summarized as follows:
\begin{itemize}
    \item We propose DPBridge, a generative framework to address general dense prediction tasks by formulating them as instances of latent diffusion bridges.
    \item We derive a tractable reverse transition kernel for diffusion bridge process, enabling maximum likelihood training scheme. This is compatible with pretrained backbones, thus improving both training stability and the exploitation of visual prior knowledge.
    \item We introduce finetuning strategies to improve finetuning performance, including a distribution-aligned normalization technique that mitigates distributional discrepancies between the bridge and standard diffusion processes, and an auxiliary image consistency loss to better preserve fine-grained spatial details.
    \item Our method shows competitive performance across different tasks in terms of both quantitative metrics and visual inspection.
\end{itemize}

\section{Related Works}
\label{related works}
\noindent\textbf{Dense Prediction.} Common solutions to dense predictions are categorized into discriminative and generative ones. Discriminative methods directly map input data to structured outputs in a feed-forward manner. \cite{ranftl2021vision} introduces transformer architecture and fusion mechanism to better capture both local and global context features. \cite{bae2024rethinking} incorporates per-pixel ray directions and models the relative rotations between neighboring pixels. Recently, diffusion-based generative methods have been applied in this field and have shown remarkable performance. \cite{ke2024repurposing} finetunes pretrained diffusion backbone, leveraging the rich visual knowledge within to enhance depth estimation. \cite{fu2025geowizard} takes advantage of diffusion priors to jointly estimate depth and surface normals from single image, enhancing detail preservation by decoupling complex data distributions into distinct sub-distributions. \cite{gui2024depthfm} exploits flow matching to facilitate the distribution shift between images and depth maps. Our work follows the conventional generative pipeline, but instead of starting from pure noise, we establish direct connections between input image and target signal map distributions, and model conditional probability transport in an iterative denoising manner.

\noindent\textbf{Diffusion and Bridge Models.} Diffusion models~\cite{ho2020denoising,song2020score} are a family of generative models that excel at capturing complex data distributions through a sequence of forward and reverse processes. They have driven the advancement of modern generative systems, and have consistently achieved state-of-the-art performance across various domains, such as image and video synthesis~\cite{rombach2022high, ho2022video}, controlled editing~\cite{zhang2023adding, batzolis2021conditional}, inverse problem solving~\cite{chung2022diffusion,song2023pseudoinverse}, 3D assets generation~\cite{xiang2024structured,zhang2024clay}, and human-centered applications~\cite{ji2024jade,ji2024unsupervised,ji20243d}.

In recent years, diffusion bridge models have gained growing interest in generative modeling~\cite{pavon2021data,liu2022let,zhou2023denoising,liu2023learning,de2021diffusion}. They offer a novel perspective for translating between distributions and enable a more effective data-to-data generation paradigm. Recent works have explored different types of bridge processes and their applications. \cite{liu20232,li2023bbdm,yue2023image} develop Schrödinger bridge, Brownian bridge, and Ornstein-Uhlenbeck bridge frameworks, applying them to image restoration tasks such as inpainting, super-resolution. \cite{chung2024direct} proposes a modified inference procedure that enforces data consistency. \cite{chen2023schrodinger} introduces bridge models to text-to-speech systems, enhancing synthesis quality. 

To the best of our knowledge, no prior work has explored the use of bridge models for dense prediction tasks to facilitate distribution transport between images and dense signal maps, despite their structural closeness. Our work aims to fill this research gap. Conceptually, DPBridge is most related to DDBM~\cite{zhou2023denoising}, but the two differ in several key aspects: DDBM is trained from scratch using a score matching objective, whereas DPBridge is finetuned from large pretrained diffusion backbones using a maximum likelihood objective. In addition, we introduce strategies for efficient finetuning, enabling better use of visual priors and significantly improving performance on dense prediction settings.

\section{Preliminary}
\label{preliminary}

\subsection{Diffusion Process}
\label{sec:diffusion preliminary}
A diffusion process $\mathbb{Q}=\{\mathbf{z}_t \in \mathbb{R}^d: t \in [0, T]\}$ on space $\mathbb{R}^d$ follows the stochastic differential equation of form
\begin{equation}
\label{eq:diffusion SDE}
    \mathrm{d}\mathbf{z}_t = f(\mathbf{z}_t, t)\mathrm{d}t + g(t)\mathrm{d}\mathbf{w}_t
\end{equation}
where $\mathbf{w}_t$ is a standard Wiener process, $f: \mathbb{R}^d \times [0, T] \rightarrow \mathbb{R}$ and $g: [0, T] \rightarrow \mathbb{R}$ are the drift and diffusion coefficients that describe the evolution of this stochastic process. The transition kernel is specified as $q(\mathbf{z}_t | \mathbf{z}_0) = \mathcal{N}(\alpha_t\mathbf{z}_0, \eta_t^2 \mathbf{I})$ and the noise schedule comprising differentiable functions $\alpha_t, \eta_t$ depends on the design choice of drift and diffusion coefficients. It is also common to assess the diffusion process in terms of the signal-to-noise ratio (SNR), defined as $\text{SNR}_t = \alpha_t^2 / \eta_t^2$.

\subsection{Diffusion Bridge Process}
\label{sec:diffusion bridge preliminary}
Diffusion bridge process is a special type of diffusion process that is conditioned to end at a specified terminal point at timestep $T$. Its dual fixed-endpoint characteristics enable a bidirectional distribution transport. Its construction can be derived from a reference diffusion process conditioned on terminal constraints. We can use Doob’s h-transform technique~\cite{oksendal2013stochastic,sarkka2019applied} to modify a reference process $\mathbb{Q}$ defined in \cref{eq:diffusion SDE} to ensure that the event $\mathbf{z}_T=\mathbf{x}$ happens, where $\mathbf{x}$ is a predefined value. The conditioned process $\mathbb{Q}^{\mathbf{x}}(\cdot):=\mathbb{Q}\left(\cdot | \mathbf{z}_T=\mathbf{x}\right)$ thus follows the law of
\begin{equation}
\label{eq:diffusion bridge SDE}
    \mathrm{d}\mathbf{z}_t = \left[ f(\mathbf{z}_t, t) + g^2(t) h(\mathbf{z}_t, t, \mathbf{x}, T)\right] \mathrm{d}t + g(t) \mathrm{d}\mathbf{w}_t
\end{equation}
where $h(\mathbf{z}_t, t, \mathbf{x}, T) = \nabla_{\mathbf{z}_t} \log q(\mathbf{z}_T=\mathbf{x} | \mathbf{z}_t)$. Intuitively, the additional drift term $h(\mathbf{z}_t, t, \mathbf{x}, T)$ plays the role of steering $\mathbf{z}_t$ towards the target point $\mathbf{z}_T=\mathbf{x}$, and it is ensured that $\mathbb{Q}^{\mathbf{x}}$ invariably passes $\mathbf{x}$ at timestep $T$ with probability 1 for any initial point $\mathbf{z}_0$.

The transition kernel of $\mathbb{Q}^{\mathbf{x}}$, given by $q(\mathbf{z}_t | \mathbf{z}_0, \mathbf{z}_T)$, can be subsequently derived through Kolmogorov equation:
\begin{equation}
\begin{aligned}
\label{eqn: bridge forward kernel}
    q(\mathbf{z}_t | \mathbf{z}_0, \mathbf{z}_T) &= \mathcal{N}(m_t\mathbf{z}_0 + n_t\mathbf{z}_T, \sigma_t^2 \mathbf{I})  \\
    m_t &= \alpha_t \left( 1 - \frac{\text{SNR}_T}{\text{SNR}_t} \right), \quad
    n_t = \frac{\text{SNR}_T}{\text{SNR}_t} \frac{\alpha_t}{\alpha_T}, \\
    \sigma_t &= \eta_t \sqrt{ 1 - \frac{\text{SNR}_T}{\text{SNR}_t}}
\end{aligned}
\end{equation}
where $\text{SNR}_t, \alpha_t, \sigma_t^2$ are all defined in the reference process $\mathbb{Q}$. In \cref{eqn: bridge forward kernel}, $m_t$ and $n_t$ are coefficients that enforce the bridge’s dual endpoints: at $t=0, m_0=1, n_0=0 \rightarrow z_t=z_0$; at $t=T, m_T=0, n_T=1 \rightarrow z_t=z_T$.

\section{Method}
\label{method}
\subsection{Problem Formulation}
We formulate different dense prediction tasks as instances of image-conditioned generation problems and propose DPBridge to model the conditional distributions $p(\mathbf{y} | \mathbf{x})$, where $\mathbf{y} \in \mathbb{R}^{H \times W \times C}$ denotes the target dense signal map (e.g. depth, surface normal, segmentation map), and $\mathbf{x} \in \mathbb{R}^{H \times W \times 3}$ is the input RGB image. In the following sections, we'll elaborate the details of the DPBridge framework, which is constructed based on pretrained diffusion backbone through proper finetuning.

\subsection{Dense Prediction Bridge}
\begin{figure*}[t]
    \begin{center}
    \includegraphics[width=\linewidth]{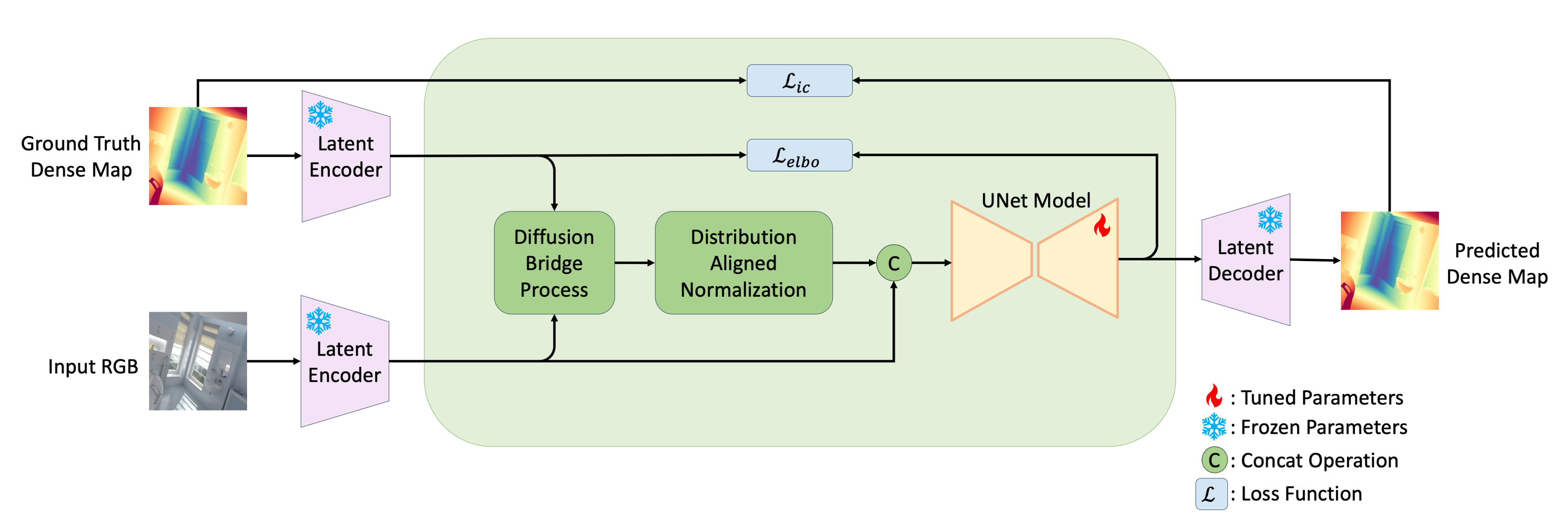}
    \caption{Overview of our training pipeline. We derive the maximum likelihood training objective of diffusion bridge process, propose a distribution-aligned normalization strategy, and assist the training with an auxiliary image consistency loss.}
    \label{fig:training pipeline}
    \end{center}
\end{figure*}

Given paired training data, a straightforward approach is to construct the diffusion bridge model from scratch, following the formulation in \cref{sec:diffusion bridge preliminary}. However, this approach is extremely resource-demanding, making it hard to implement with limited computation power. What's more, it cannot to harness rich visual prior embedded in the foundation models pretrained with large-scale data, which have been shown to significantly enhance both perceptual quality and generalization capability in dense prediction tasks. To address these limitations and fully exploit the visual prior, our DPBridge is built upon pretrained diffusion backbones through finetuning. We explore several strategies to handle the misalignment between bridge process and standard diffusion process, facilitating more efficient knowledge transfer. An overview of our training pipeline is provided in \cref{fig:training pipeline}, followed by a detailed explanation of each component.

\subsection{Latent Space Implementation}
We follow common practice to perform diffusion steps in a compressed yet perceptually aligned latent space, provided by an independently trained variational autoencoder (VAE). Specifically, both the dense signal map and RGB image are encoded using the VAE encoder, yielding latent representations $\mathbf{z}_0=\mathcal{E}(\mathbf{y})$ and $\mathbf{z}_T=\mathcal{E}(\mathbf{x})$, which serve as the initial and terminal points of the diffusion bridge process. In this way the diffusion bridge process is operated entirely in the latent space. To recover the original signal map, we use the VAE decoder to decode the predicted latent $\hat{\mathbf{y}}=\mathcal{D}(\hat{\mathbf{z}}_0)$. For clarity, all subsequent derivations in this paper are conducted in latent space.

\subsection{Training Objective}
\label{sec:train loss}
\textbf{Maximum Likelihood Loss.} A widely adopted approach for learning the reverse process associated with \cref{eq:diffusion bridge SDE} follows the theoretical framework of~\cite{anderson1982reverse}, where a neural network is trained to approximate the score function~\cite{zhou2023denoising}. In contrast, we observe that in the diffusion bridge setting, the reverse transition kernel admits a tractable analytical form. This insight enables us to adopt a maximum likelihood training paradigm in place of score matching. Since widely used latent diffusion models like Stable Diffusion~\cite{rombach2022high} are pretrained under a likelihood-based loss function, finetuning with a consistent objective is more compatible, which helps preserve prior knowledge and improves training stability.

Specifically, the training process is performed by optimizing the Evidence Lower Bound (ELBO). Given the encoded training pair $\big(\mathbf{z}_0, \mathbf{z}_T\big)$, the expectation of data log-likelihood possesses the following ELBO formulation:
\begin{equation}
    \begin{aligned}
        \emph{ELBO} = &\mathbb{E}_{p(\mathbf{z}_0)} \Big[ \mathbb{E}_{p(\mathbf{z}_1 \mid \mathbf{z}_0)} \big[ \log p_\theta (\mathbf{z}_0 \mid \mathbf{z}_1, \mathbf{z}_T) \big] - \\
        \sum_{t=2}^{T} \mathbb{E}_{p(\mathbf{z}_t | \mathbf{z}_0)} &\big[ \mathbb{D}_{KL} \big( q(\mathbf{z}_{t-1} \mid  \mathbf{z}_t, \mathbf{z}_0, \mathbf{z}_T) \,||\, p_\theta(\mathbf{z}_{t-1} \mid \mathbf{z}_t, \mathbf{z}_T) \big) \big] \Big]
    \end{aligned}
\end{equation}
where $q(\mathbf{z}_{t-1} | \mathbf{z}_t, \mathbf{z}_0, \mathbf{z}_T)$ and $p_\theta(\mathbf{z}_{t-1} | \mathbf{z}_t, \mathbf{z}_T)$ are the transition kernel of true reverse process and our modeled generative process. $\mathbb{D}_{KL}$ is the KL divergence. Utilizing Bayes’ theorem and Markov chain property, we can obtain:
\begin{equation}
\begin{aligned}
\label{eqn: bridge reverse kernel}
    q(\mathbf{z}_{t-1} | \mathbf{z}_t, \mathbf{z}_0, \mathbf{z}_T) &= \frac{q(\mathbf{z}_t | \mathbf{z}_{t-1}, \mathbf{z}_T)q(\mathbf{z}_{t-1} | \mathbf{z}_0, \mathbf{z}_T)}{q(\mathbf{z}_t | \mathbf{z}_0, \mathbf{z}_T)}\\
    &= \mathcal{N}(\tilde{\boldsymbol{\mu}}_t(\mathbf{z}_t, \mathbf{z}_0, \mathbf{z}_T), \tilde{\sigma}_t^2\mathbf{I})  \\
\end{aligned}
\end{equation}
For simplicity, we define the following notations: $a_t = m_t / m_{t-1}, b_t = n_t - a_t \cdot n_{t-1}, \delta_t^2 = \sigma_t^2 - a_t^2 \cdot \sigma_{t-1}^2$. Based on \cref{eqn: bridge forward kernel}, we further derive the mean value term:
\begin{equation}
\begin{aligned}
\label{eqn: bridge reverse mean}
    \tilde{\boldsymbol{\mu}}_t(\mathbf{z}_t, \mathbf{z}_0, \mathbf{z}_T) &= k_1 \mathbf{z}_t + k_2 \mathbf{z}_0 + k_3 \mathbf{z}_T \\
    k_1 &= \frac{\sigma_{t-1}^2}{\sigma_t^2} \cdot a_t, \quad
    k_2 = \frac{\delta_t^2}{\sigma_t^2} \cdot m_{t-1}, \\
    k_3 &= \frac{\delta_t^2}{\sigma_t^2} \cdot n_{t-1} - \frac{\sigma_{t-1}^2}{\sigma_t^2} \cdot a_t \cdot b_t,
\end{aligned}
\end{equation}
and the variance term:
\begin{equation}
\begin{aligned}
\label{eqn: bridge reverse variance}
    \tilde{\sigma}_t^2 = \frac{\sigma_{t-1}^2}{\sigma_t^2} \cdot \delta_t^2
\end{aligned}
\end{equation}

Assuming $p_\theta(\mathbf{z}_{t-1} | \mathbf{z}_t, \mathbf{z}_T)$ also follows a Gaussian distribution $\mathcal{N}(\boldsymbol{\mu}_{\theta}(\mathbf{z}_t, \mathbf{z}_T, t), \boldsymbol{\Sigma}_{\theta}(\mathbf{z}_t, \mathbf{z}_T, t))$ with fixed variance $\boldsymbol{\Sigma}_{\theta}(\mathbf{z}_t, \mathbf{z}_T, t)=\tilde{\sigma}_t^2$ for simplicity of derivation. Following the strategy of DDPM~\cite{ho2020denoising}, we reparameterize the model $\boldsymbol{\epsilon}_{\theta}(\mathbf{z}_t, \mathbf{z}_T, t)$ to predict noise instead of posterior mean. Under this formulation, the ELBO objective is equivalent to:
\begin{equation}
\begin{aligned}
\label{eqn: diffusion loss}
        \mathcal{L}_{elbo} &= \mathbb{E}_{t, \mathbf{z}_0, \mathbf{z}_t, \mathbf{z}_T} \left[ \frac{1}{2 \tilde{\sigma}_t^2} \left\| \tilde{\boldsymbol{\mu}}_t - \boldsymbol{\mu}_{\theta}(\mathbf{z}_t, \mathbf{z}_T, t) \right\|^2 \right] \\
        &\cong \mathbb{E}_{t, \mathbf{z}_0, \mathbf{z}_t, \mathbf{z}_T} \left[ \left\| \boldsymbol{\epsilon} - \boldsymbol{\epsilon}_{\theta}(\mathbf{z}_t, \mathbf{z}_T, t) \right\|^2 \right]
\end{aligned}
\end{equation}
While \cref{eqn: diffusion loss}’s MSE-like form resembles DDPM’s noise loss, its derivation relies on the analytical reverse kernel of the bridge process. The inclusion of $z_T$ is not “standard conditional generation” but a requirement for enforcing the bridge’s dual-endpoint constraint. For more derivation details, please refer to \cref{app:derivation details}

\noindent\textbf{Image Consistency Loss.} In addition to the latent-space likelihood loss, we also incorporate an image-level consistency loss, where we first decode the latent prediction back to image space, and then evaluate its consistency with the ground truth dense map using a simple MSE.
\begin{equation}
    \mathcal{L}_{ic} = \left\| \mathcal{D}(\hat{\mathbf{z}}_0) - \mathbf{y}) \right\|^2
\end{equation}
This additional supervision addresses the loss of high-frequency structures caused by latent-space compression, which can impair the model’s ability to preserve fine-grained details. By introducing this auxiliary image-space loss, we encourage the model to retain spatial information critical for accurate per-pixel prediction. This dual-loss strategy combines the efficiency of latent-space modeling with the fidelity of image-space supervision, leading to improved performance. The final training objective is given by:
\begin{equation}
    \mathcal{L} = \omega_1\mathcal{L}_{elbo} + \omega_2\mathcal{L}_{ic} 
\end{equation}
where $\omega_1, \omega_2$ are hyperparameters to balance the loss.

\subsection{Distribution Alignment Normalization}
A key challenge during finetuning is the misalignment between the marginal distribution $q(\mathbf{z}_t)$ of intermediate samples in DPBridge and that of the diffusion backbone, due to the difference in the forward transition kernels. This misalignment weakens the model's denoising ability and hinders effective knowledge transfer. To address this issue, we propose a distribution alignment normalization technique that transforms intermediate latents computed from our bridge process \cref{eqn: bridge forward kernel} to match the distributional form of a standard diffusion process.
\begin{equation}
\label{eqn:normalization}
    \mathbf{z}_t' = \frac{\mathbf{z}_t - m_t \mathbf{z}_T}{\sqrt{n_t^2 + \sigma_t^2}} = \frac{n_t}{\sqrt{n_t^2 + \sigma_t^2}} \mathbf{z}_0 + \frac{\sigma_t}{\sqrt{n_t^2 + \sigma_t^2}} \boldsymbol{\epsilon}_t
\end{equation}
Since the square sum of coefficients is equal to 1, the marginal distribution of the normalized bridge process $\{\mathbf{z}_t': t \in [0, T]\}$ can thus align with that of the standard diffusion process, while preserving bridge-specific dynamics. Therefore, by feeding the normalized sample $\mathbf{z}_t'$ in the network $\boldsymbol{\epsilon}_{\theta}$, we can fully exploit the denoising capability from the pretrained diffusion backbone. We note that a similar strategy was independently proposed in FrameBridge~\cite{wang2024framebridge}, where it is applied in the context of image-to-video generation. Our method, developed concurrently, is specifically designed to facilitate adaptation in dense prediction tasks under the DPBridge framework. An overview of the full training and inference procedures is summarized in \cref{alg:training} \cref{alg:inference}.

\section{Experiments}
We conduct evaluation primarily on two representative dense prediction tasks: depth estimation and surface normal prediction. We begin with a brief overview of experimental settings, followed by both quantitative and qualitative analyses. Next, we present ablation studies to investigate the contribution of each design component. Implementation details are presented in \cref{app:implem details} for reference.

\subsection{Experiments Setup}
\label{sec:exp setup}
\noindent\textbf{Datasets.} The basic setting for depth estimation and surface normal prediction experiments are similar. We use two synthetic datasets, HyperSim~\cite{roberts2021hypersim} and Virtual KITTI~\cite{cabon2020virtual}, for training, which cover both indoor and outdoor scenes with high-quality ground-truth annotations. To comprehensively evaluate the model performance and demonstrate its generalization capability, we conduct zero-shot evaluations on real-world datasets. For depth estimation, this includes NYUv2~\cite{silberman2012indoor}, KITTI~\cite{geiger2012we}, ScanNet~\cite{dai2017scannet}, ETH3D~\cite{schops2017multi} and DIODE~\cite{vasiljevic2019diode}. For surface normal prediction, we additionally incorporate iBims~\cite{koch2018evaluation} and Sintel~\cite{butler2012naturalistic} datasets. For fair comparison, we use the same validation split as in previous methods~\cite{ke2024repurposing}.

\noindent\textbf{Evaluation Metrics.} For depth estimation, we follow the affine-invariant evaluation protocol, where we first align the prediction results to the ground truth with least squares fitting, and then use Absolute Mean Relative Error (AbsRel) and $\delta_1$ accuracy to assess the performance. Denote the ground truth depth $\mathbf{d}$, predicted depth $\hat{\mathbf{d}}$, and the number of valid pixels $N$, the AbsRel is calculated as $\frac{1}{N}\sum_{i=1}^N\|\mathbf{d}_i-\hat{\mathbf{d}}_i\| / \mathbf{d}_i$, and $\delta_1$ accuracy is measured using the ratio
of pixels satisfying $\text{max}(\mathbf{d}_i/\hat{\mathbf{d}}_i, \hat{\mathbf{d}}_i/\mathbf{d}_i) < 1.25$. For the metrics of surface normal task, we report the commonly used mean angular error between the ground-truth normal vectors and the prediction results, as well as the percentage of pixels with an angular error below 11.25 degrees.

\begin{table*}[t]
\centering
\resizebox{\linewidth}{!}{%
\begin{tabular}{lcccccccccccccc}
\toprule
\textbf{Method} & \multicolumn{3}{c}{\textbf{Model Type}} & \multicolumn{2}{c}{\textbf{NYUv2}} & \multicolumn{2}{c}{\textbf{KITTI}} & \multicolumn{2}{c}{\textbf{ETH3D}} & \multicolumn{2}{c}{\textbf{ScanNet}} & \multicolumn{2}{c}{\textbf{DIODE}} & \textbf{Avg.}\\ 
& FF & DM & BM & AbsRel$\downarrow$ & $\delta_1\uparrow$ & AbsRel$\downarrow$ & $\delta_1\uparrow$ & AbsRel$\downarrow$ & $\delta_1\uparrow$ & AbsRel$\downarrow$ & $\delta_1\uparrow$ & AbsRel$\downarrow$ & $\delta_1\uparrow$ & \textbf{Rank}\\
\midrule
DiverseDepth~\cite{yin2020diversedepth} & \checkmark & &  & 11.7 & 87.5 & 19.0 & 70.4 & 22.8 & 69.4 & 10.9 & 88.2 & 37.6 & 63.1 & 12.2 \\
MiDaS~\cite{ranftl2020towards} & \checkmark & & & 11.1 & 88.5 & 23.6 & 63.0 & 18.4 & 75.2 & 12.1 & 84.6 & 33.2 & 71.5 & 11.8 \\
LeReS~\cite{yin2021learning} & \checkmark & & & 9.0 & 91.6 & 14.9 & 78.4 & 17.1 & 77.7 & 9.1 & 91.1 & 27.1 & 76.6 & 9.5 \\
Omnidata~\cite{eftekhar2021omnidata} & \checkmark & & & 7.4 & 94.5 & 14.9 & 83.5 & 16.6 & 77.8 & 7.5 & 93.2 & 33.9 & 74.2 & 9.2 \\
HDN~\cite{zhang2022hierarchical} & \checkmark & & & 6.9 & 94.8 & 11.5 & 86.7 & 12.1 & 83.3 & 8.0 & 93.6 & 24.6 & 78.0 & 7.2 \\
DPT~\cite{ranftl2021vision} & \checkmark & & & 9.8 & 90.3 & 10.0 & 90.1 & 7.8 & 94.6 & 8.2 & 93.8 & 18.2 & 75.8 & 7.0 \\
Depth Anything V2~\cite{yang2024depth} & \checkmark & & & \textbf{4.4} & \textbf{97.9} & \textbf{7.5} & \textbf{94.8} & 13.2 & 84.2 & \textbf{4.4} & \textbf{98.0} & \textbf{6.5} & \textbf{95.4} & \textbf{2.3} \\
\midrule
Marigold~\cite{ke2024repurposing} & & \checkmark & & 6.0 & 95.9 & 10.5 & 90.4 & 7.1 & \underline{95.1} & 6.9 & 94.5 & 31.0 & 71.2 & 6.0 \\
DepthFM~\cite{gui2024depthfm} & & \checkmark & & 6.5 & 95.6 & \underline{8.3} & \underline{93.4} & - & - & - & - & \underline{22.5} & 80.0 & 3.8 \\
Geowizard~\cite{fu2025geowizard}  & & \checkmark & & 5.7 & 96.2 & 14.4 & 82.0 & 7.5 & 94.3 & 6.1 & 95.8 & 31.4 & 77.1 & 6.2 \\
E2E-FT~\cite{garcia2024fine} & & \checkmark & & 5.4 & 96.4 & 9.6 & 92.1 & \textbf{6.4} & \textbf{95.9} & 5.8 & 96.5 & 30.3 & 77.6 & 3.6 \\
\midrule
I2SB~\cite{liu20232} & & & \checkmark & 23.5 & 63.6 & 35.1 & 46.4 & 53.4 & 40.5 & 21.1 & 67.4 & 55.5 & 58.2 & 14.0 \\
BBDM~\cite{li2023bbdm} & & & \checkmark & 33.2 & 49.6 & 58.6 & 25.6 & 52.0 & 40.6 & 25.2 & 61.2 & 70.1 & 46.9 & 15.2\\
DDBM~\cite{zhou2023denoising} & & & \checkmark & 26.4 & 62.2 & 55.1 & 27.6 & 69.5 & 33.8 & 23.2 & 65.6 & 52.3 & 56.9 & 14.5 \\
\midrule
\rowcolor[gray]{0.9}
\textbf{DPBridge (SD1.5)} & & & \checkmark & 6.9 & 95.7 & 11.2 & 87.3 & 8.1 & 93.2 & 7.5 & 94.7 & 29.7 & 78.6 & 5.4 \\
\rowcolor[gray]{0.9}
\textbf{DPBridge (SD2.1)} & & & \checkmark & \textbf{4.4} & \underline{97.6} & 9.8 & 91.6 & \underline{6.9} & 94.9 & \underline{4.5} & \underline{97.4} & 23.8 & \underline{84.0} & \underline{2.7} \\
\bottomrule
\end{tabular}
}
\caption{Quantitative comparison on affine-invariant depth estimation benchmarks. All metrics are presented in percentage terms. FF, DM, and BM correspond to feed-forward, diffusion models, and bridge models. The best and second-best results are highlighted in \textbf{bold} and \underline{underline} formats.}
\label{tab:depth estimation}
\end{table*}

\begin{table*}[t]
\centering
\resizebox{\linewidth}{!}{%
\begin{tabular}{lcccccccccccc}
\toprule
\textbf{Method} & \multicolumn{3}{c}{\textbf{Model Type}} & \multicolumn{2}{c}{\textbf{NYUv2}} & \multicolumn{2}{c}{\textbf{ScanNet}} & \multicolumn{2}{c}{\textbf{iBims-1}} & \multicolumn{2}{c}{\textbf{Sintel}} & \textbf{Avg.}\\ 
& FF & DM & BM & Mean$\downarrow$ & $11.25^{\circ}\uparrow$ & Mean$\downarrow$ & $11.25^{\circ}\uparrow$  & Mean$\downarrow$ & $11.25^{\circ}\uparrow$ & Mean$\downarrow$ & $11.25^{\circ}\uparrow$ & \textbf{Rank}\\
\midrule
EESNU~\cite{bae2021estimating} & \checkmark & & & \underline{16.2} & 58.6 & - & - & 20.0 & 58.5 & 42.1 & 11.5 & 6.8 \\
Omnidata v1~\cite{eftekhar2021omnidata} & \checkmark & & & 23.1 & 45.8 & 22.9 & 47.4 & 19.0 & 62.1 & 41.5 & 11.4 & 8.8 \\
Omnidata v2~\cite{kar20223d} & \checkmark & & & 17.2 & 55.5 & 16.2 & 60.2 & 18.2 & 63.9 & 40.5 & 14.7 & 6.2 \\
DSINE~\cite{bae2024rethinking} & \checkmark & & & 16.4 & 59.6 & 16.2 & 61.0 & 17.1 & 67.4 & 34.9 & 21.5 & 3.5 \\
\midrule
Marigold~\cite{ke2024repurposing} & & \checkmark & & 18.8 & 55.9 & 17.7 & 58.8 & 18.4 & 64.3 & 39.1 & 14.9 & 6.4 \\
Geowizard~\cite{fu2025geowizard} & & \checkmark & & 17.0 & 56.5 & 15.4 & 61.6 & \textbf{13.0} & 65.3 & 40.4 & 13.2 & 4.2 \\
E2E-FT~\cite{garcia2024fine} & & \checkmark &  & 16.5 & \underline{60.4} & \underline{14.7} & \underline{66.1} & \underline{16.1} & \textbf{69.7} & \underline{33.5} & \underline{22.3} & \underline{2.1} \\
StableNormal~\cite{ye2024stablenormal} & & \checkmark &  & 17.7 & 55.9 & 15.8 & 61.2 & 17.2 & 66.6 & - & - & 5.0 \\
\midrule
\rowcolor[gray]{0.9}
\textbf{DPBridge (SD1.5)} & & & \checkmark & 18.1 & 54.2 & 18.0 & 54.5 & 24.3 & 51.2 & 41.2& 19.4 & 8.0 \\
\rowcolor[gray]{0.9}
\textbf{DPBridge (SD2.1)} & & & \checkmark & \textbf{14.8} & \textbf{65.3} & \textbf{13.4} & \textbf{68.5} & 17.1 & \underline{68.0} & \textbf{25.0} & \textbf{46.3} & \textbf{1.4} \\
\bottomrule
\end{tabular}
}
\caption{Quantitative comparison on surface normal prediction benchmarks. All metrics are presented in percentage terms. FF, DM, and BM correspond to feed-forward, diffusion models, and bridge models. The best and second-best results are highlighted in \textbf{bold} and \underline{underline} formats.}
\label{tab:surface normal prediction}
\end{table*}

\begin{figure*}[!t]
  \begin{subfigure}{\linewidth}
  \centering
    \begin{minipage}[]{0.185\linewidth}\centering \fontsize{8pt}{9.6pt}\selectfont
        Input Image        
    \end{minipage}\hfill
    \begin{minipage}[]{0.185\linewidth}\centering \fontsize{8pt}{9.6pt}\selectfont
        DPBridge
    \end{minipage}\hfill
    \begin{minipage}[]{0.185\linewidth}\centering \fontsize{8pt}{9.6pt}\selectfont
        Marigold
    \end{minipage}\hfill
    \begin{minipage}[]{0.185\linewidth}\centering \fontsize{8pt}{9.6pt}\selectfont
        Geowizard
    \end{minipage}\hfill
    \begin{minipage}[]{0.185\linewidth}\centering \fontsize{8pt}{9.6pt}\selectfont
        DepthFM
    \end{minipage}\vfill %

    \begin{subfigure}[]{0.185\linewidth}\centering
        \includegraphics[width=\linewidth]{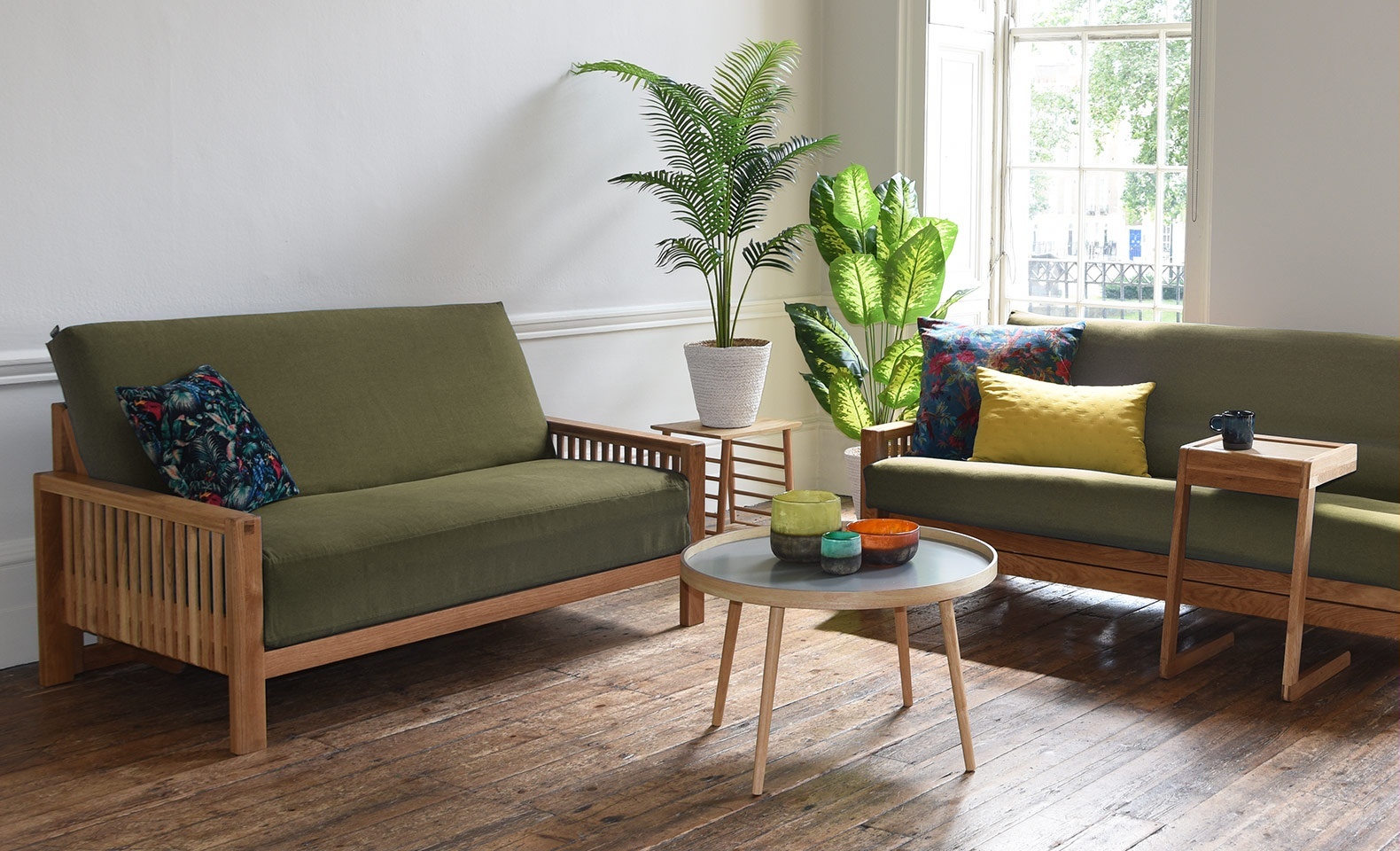}
    \end{subfigure}\hfill
    \begin{subfigure}[]{0.185\linewidth}\centering
        \includegraphics[width=\linewidth]{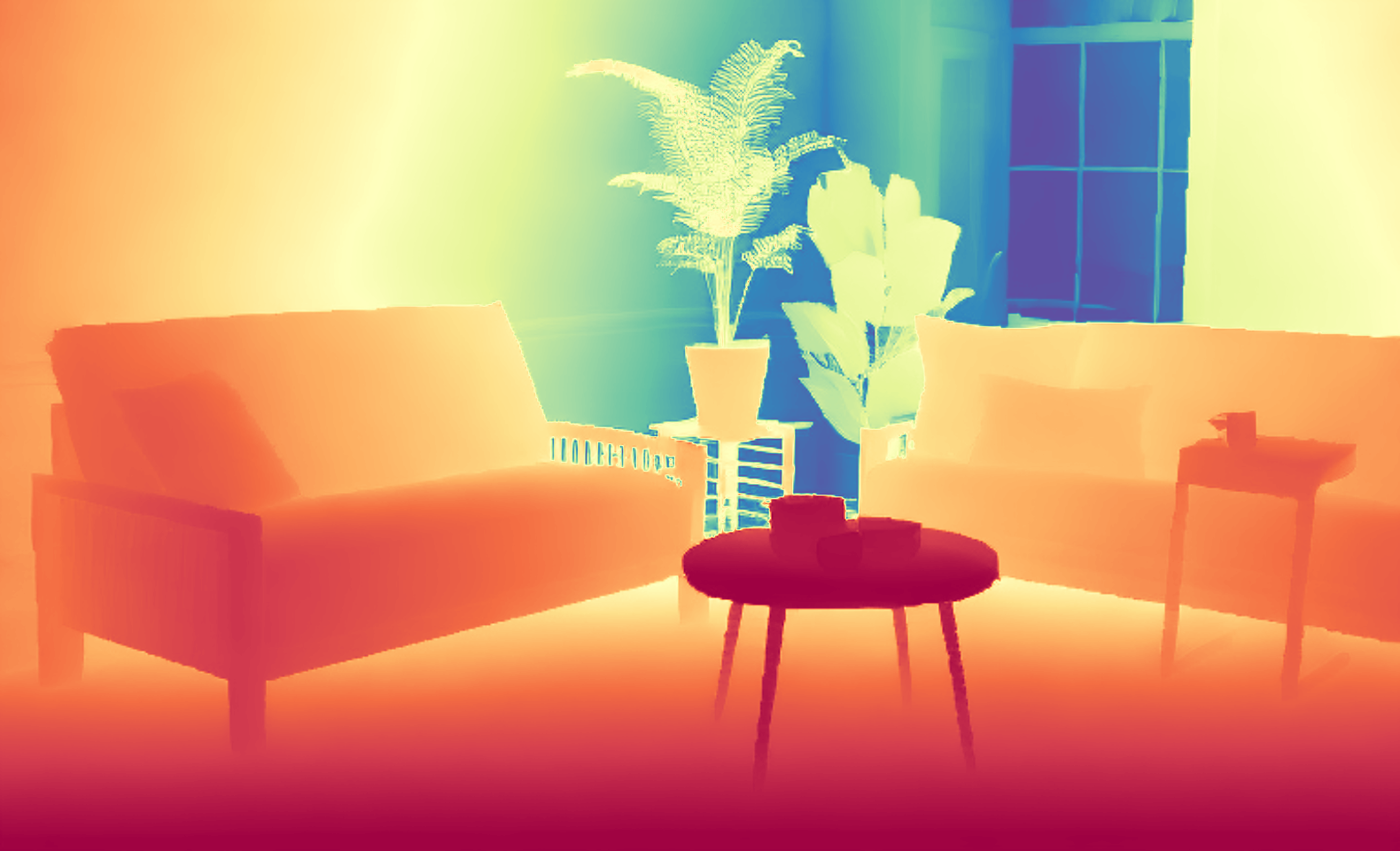}
    \end{subfigure}\hfill 
    \begin{subfigure}[]{0.185\linewidth}\centering
        \includegraphics[width=\linewidth]{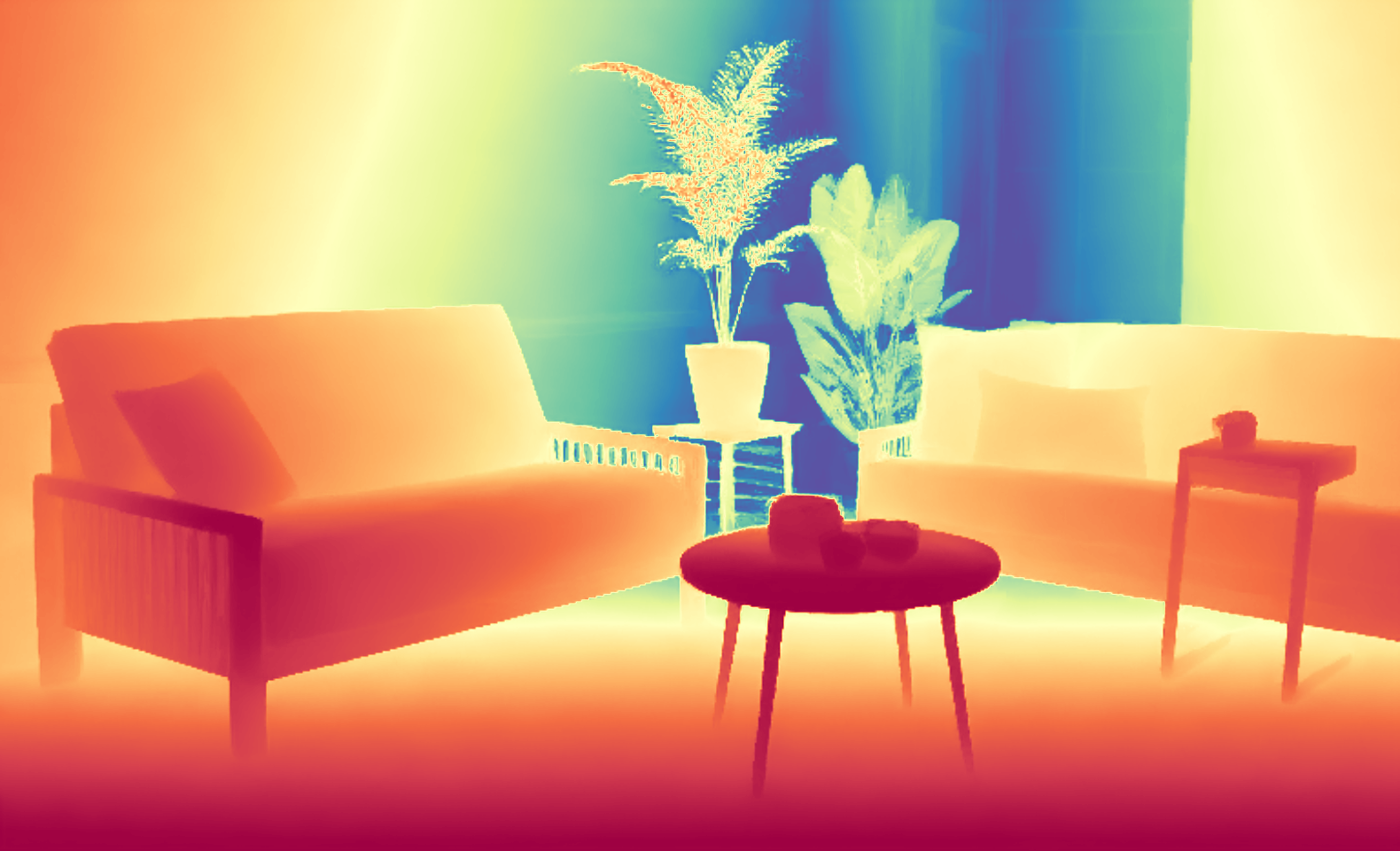}
    \end{subfigure}\hfill 
    \begin{subfigure}[]{0.185\linewidth}\centering
        \includegraphics[width=\linewidth]{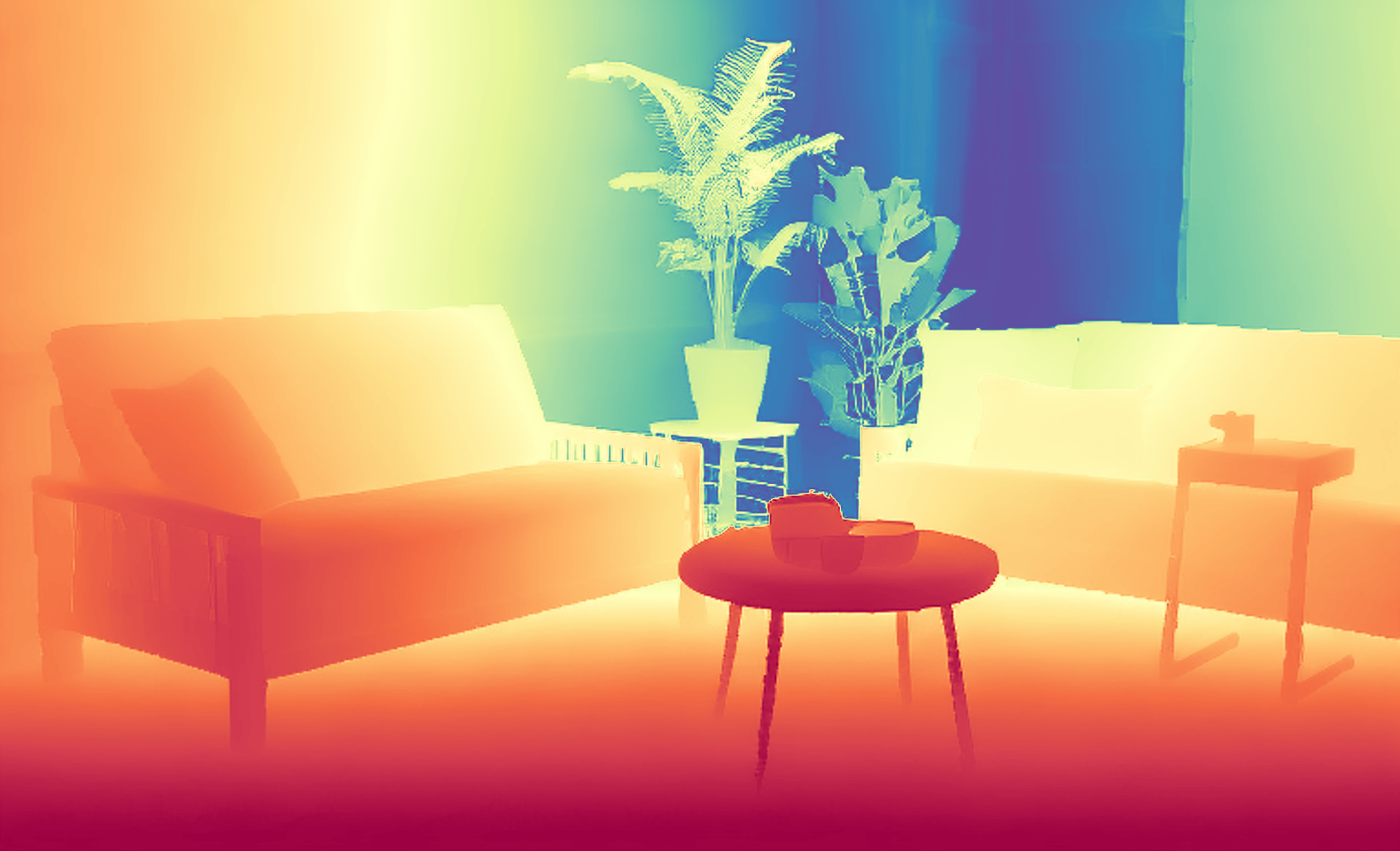}
    \end{subfigure}\hfill 
    \begin{subfigure}[]{0.185\linewidth}\centering
        \includegraphics[width=\linewidth]{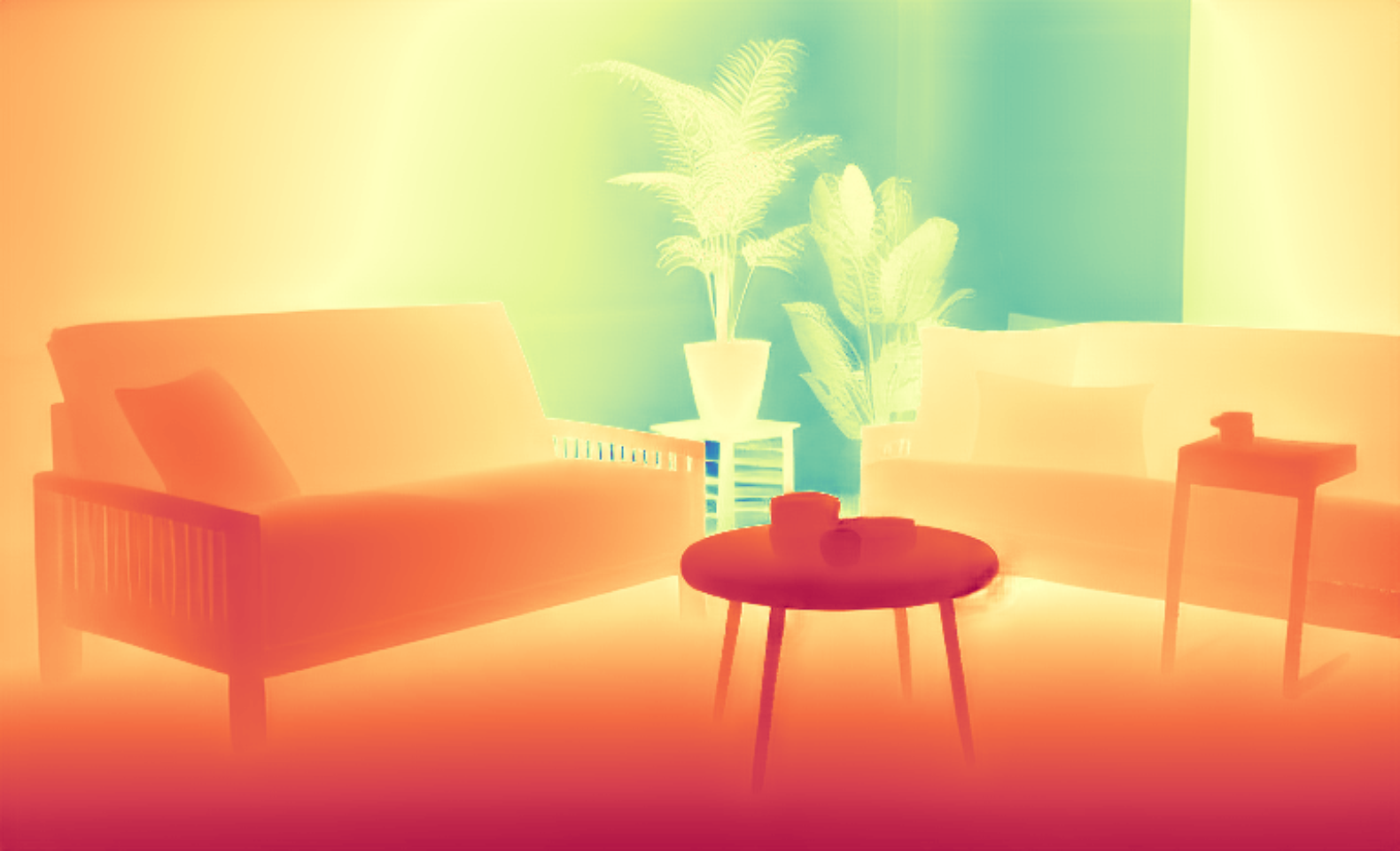}
    \end{subfigure}\vfill

    \begin{subfigure}[]{0.185\linewidth}\centering
        \includegraphics[width=\linewidth]{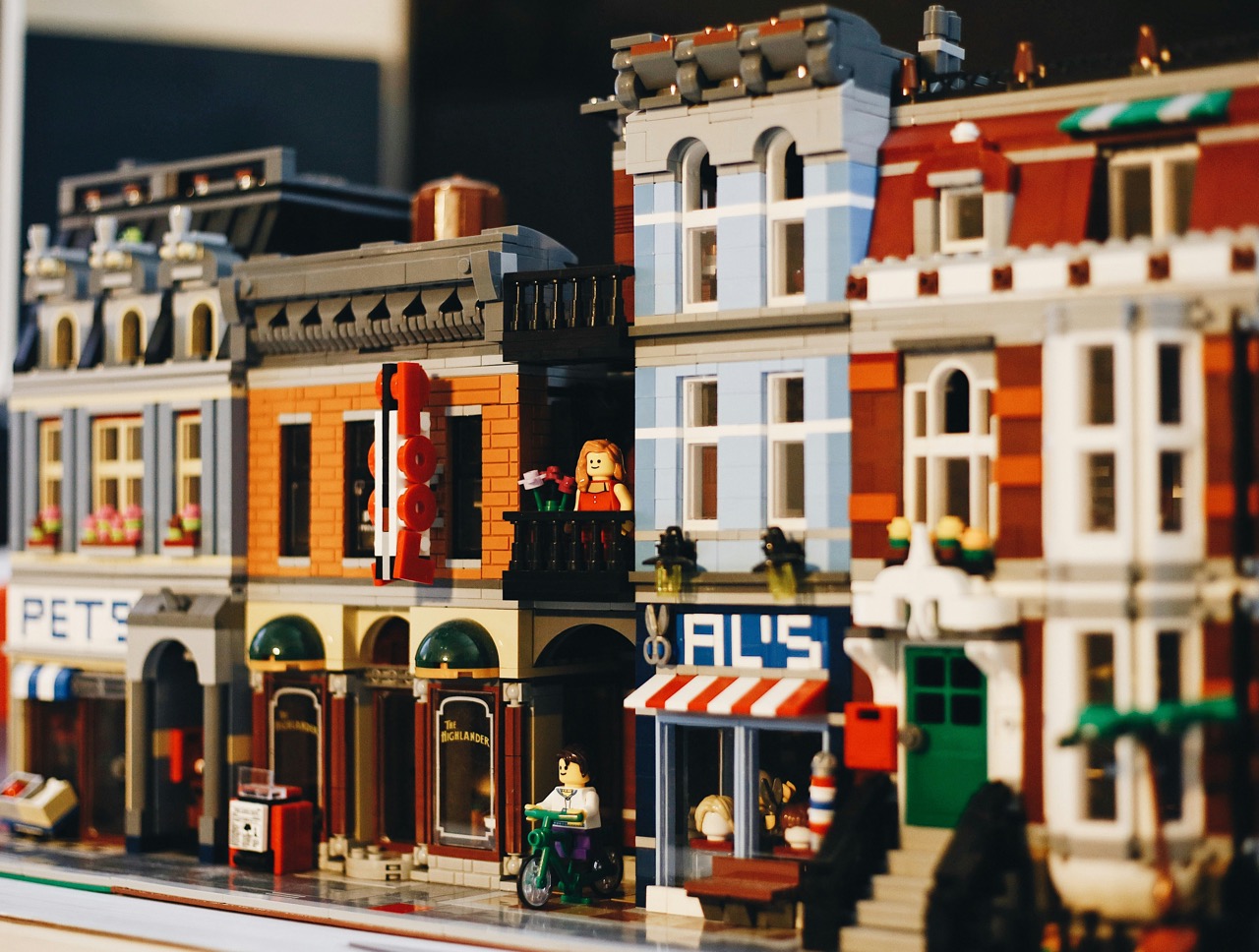}
    \end{subfigure}\hfill
    \begin{subfigure}[]{0.185\linewidth}\centering
        \includegraphics[width=\linewidth]{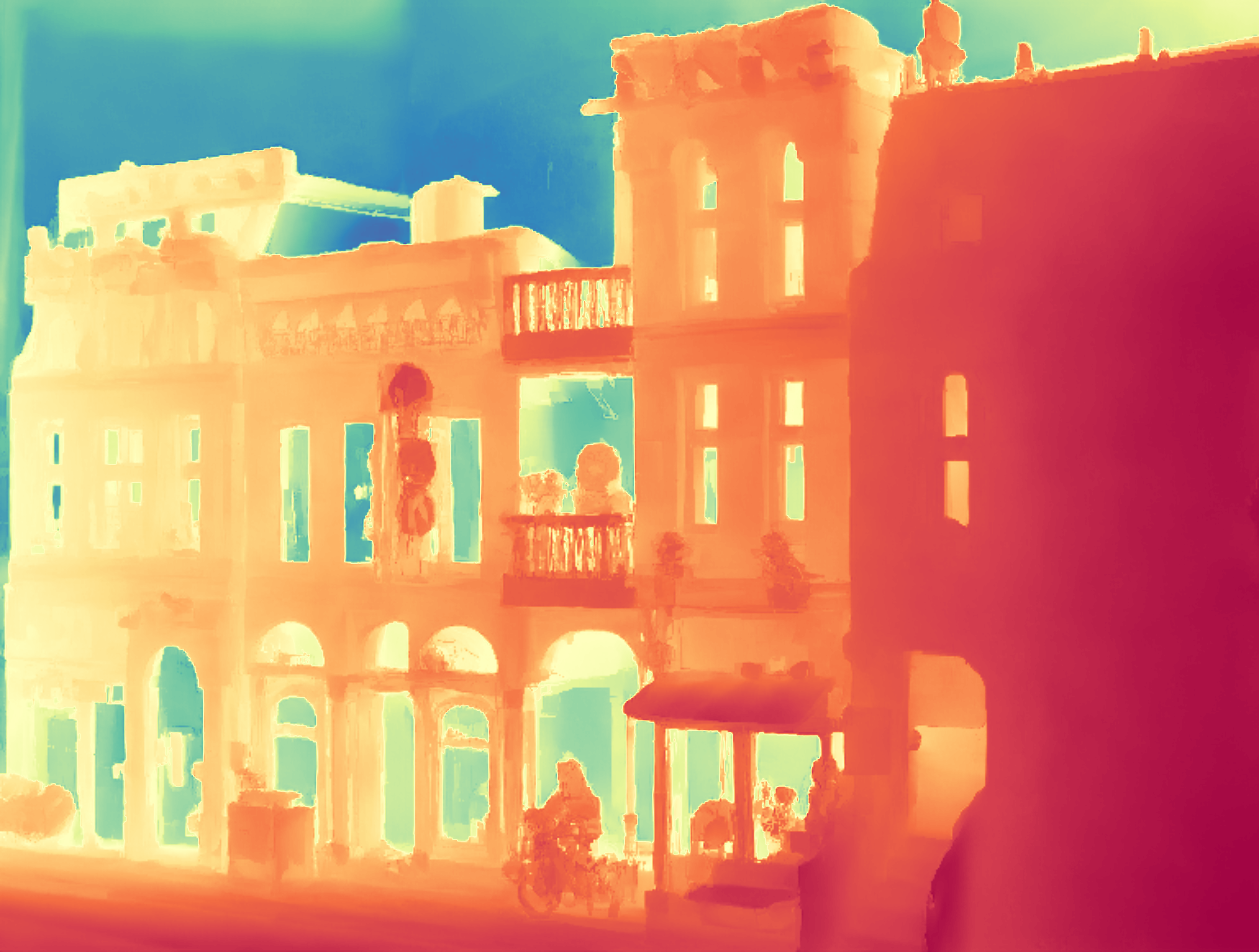}
    \end{subfigure}\hfill 
    \begin{subfigure}[]{0.185\linewidth}\centering
        \includegraphics[width=\linewidth]{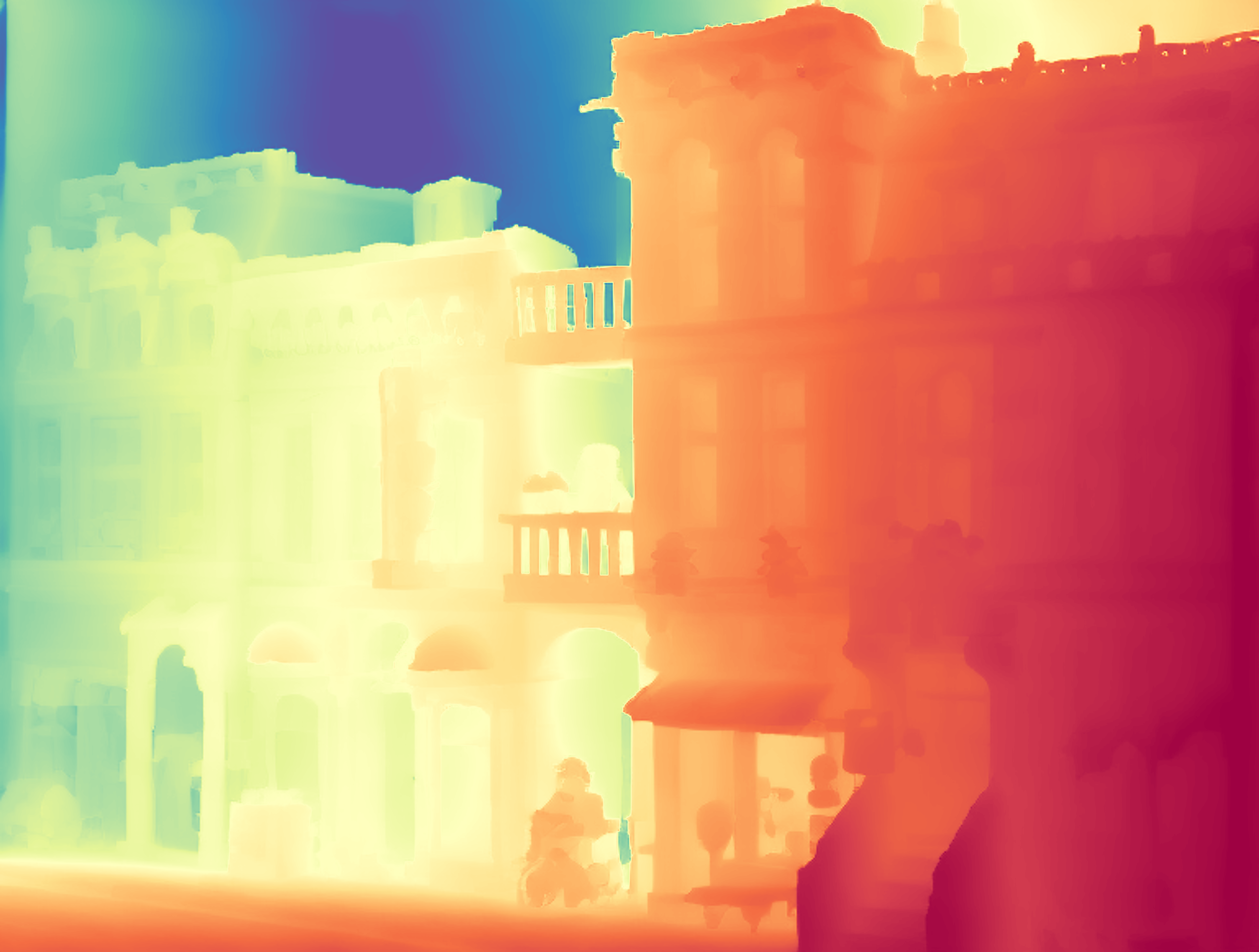}
    \end{subfigure}\hfill 
    \begin{subfigure}[]{0.185\linewidth}\centering
        \includegraphics[width=\linewidth]{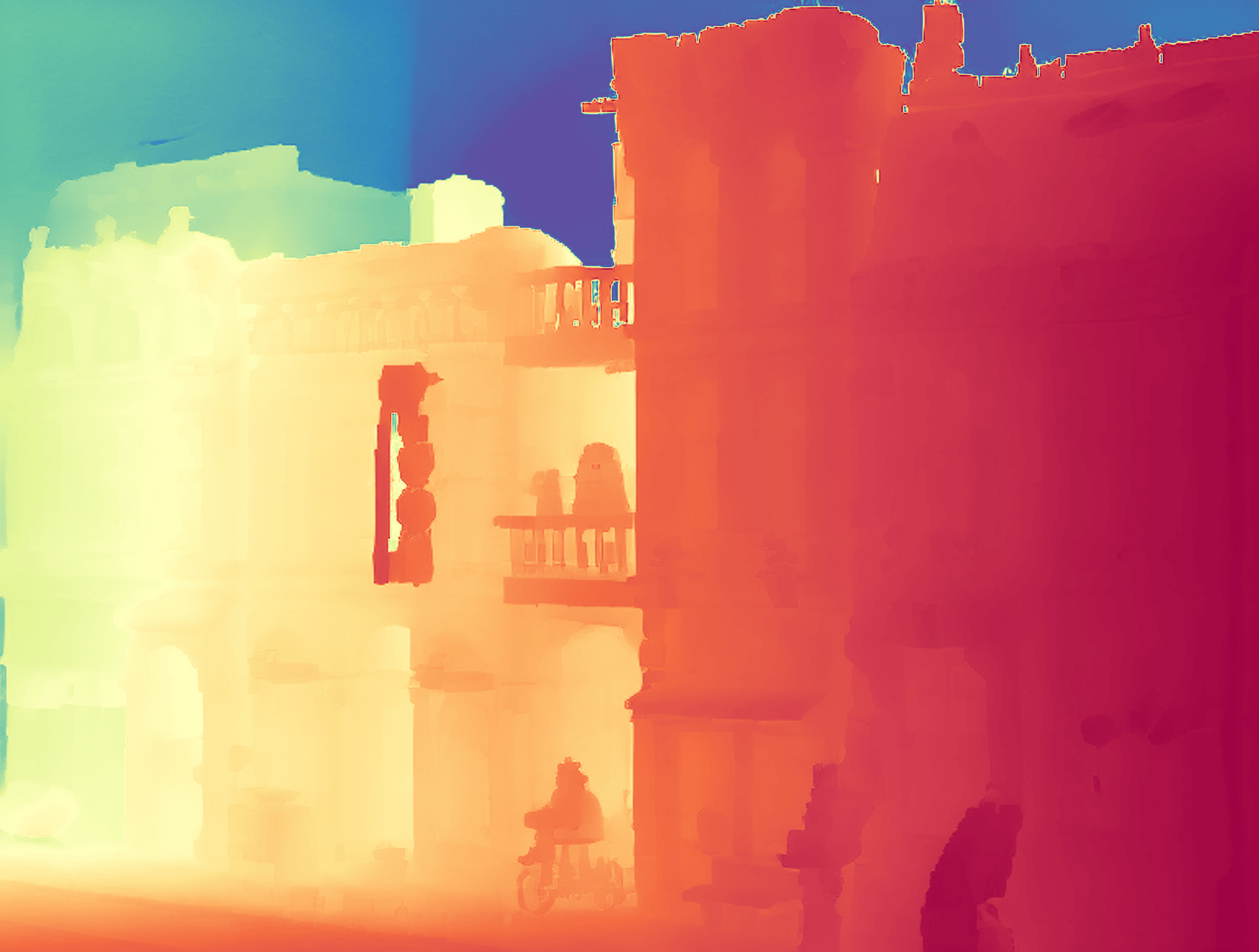}
    \end{subfigure}\hfill 
    \begin{subfigure}[]{0.185\linewidth}\centering
        \includegraphics[width=\linewidth]{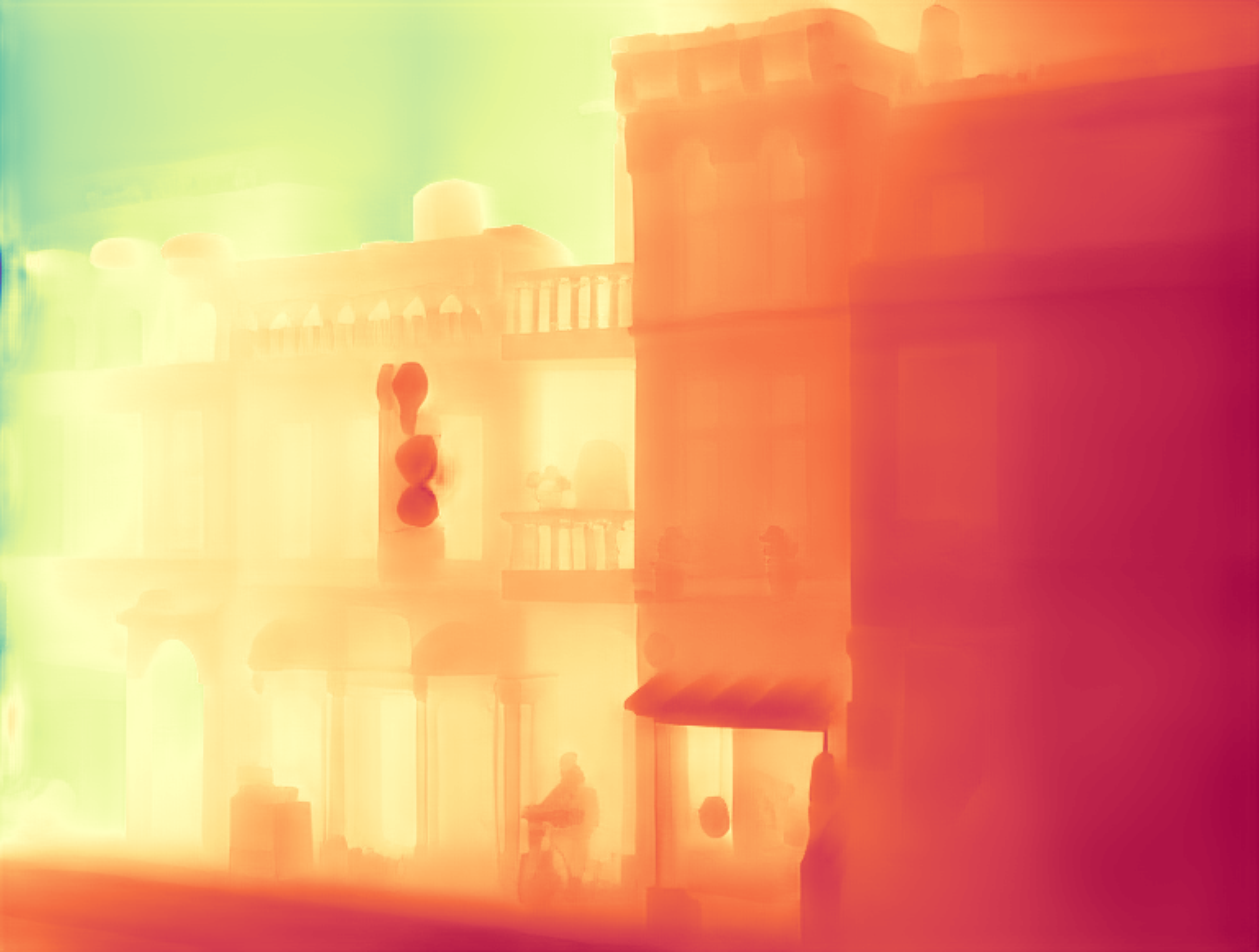}
    \end{subfigure}\vfill

    \begin{subfigure}[]{0.185\linewidth}\centering
        \includegraphics[width=\linewidth]{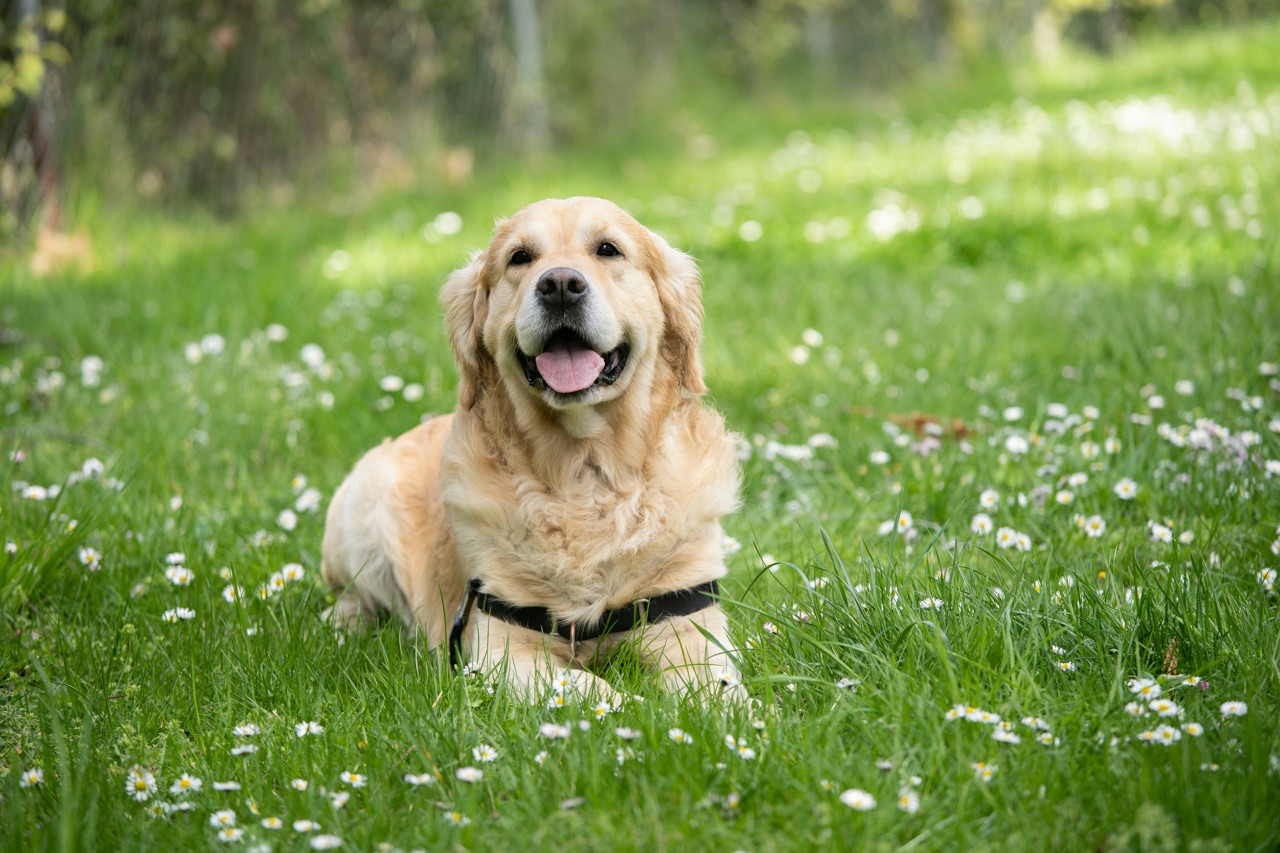}
    \end{subfigure}\hfill
    \begin{subfigure}[]{0.185\linewidth}\centering
        \includegraphics[width=\linewidth]{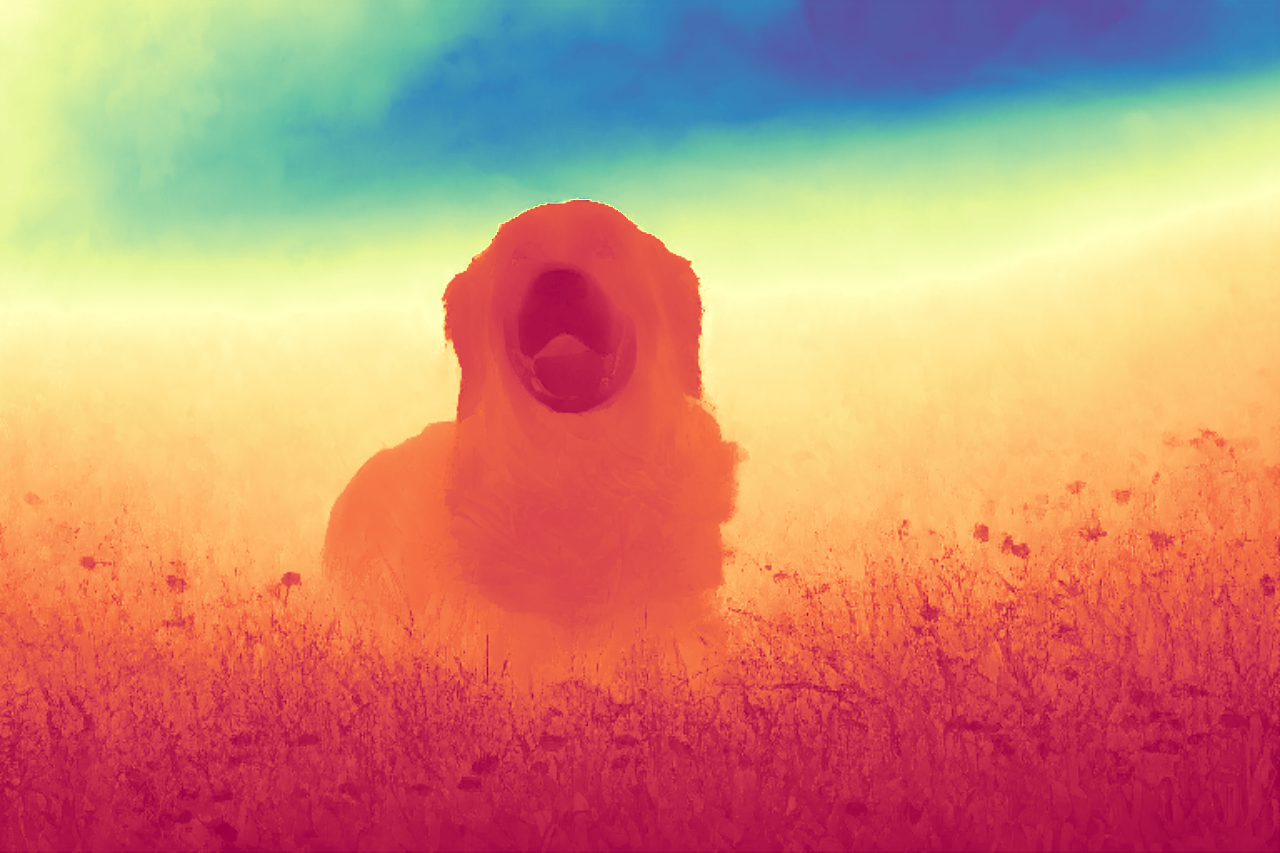}
    \end{subfigure}\hfill 
    \begin{subfigure}[]{0.185\linewidth}\centering
        \includegraphics[width=\linewidth]{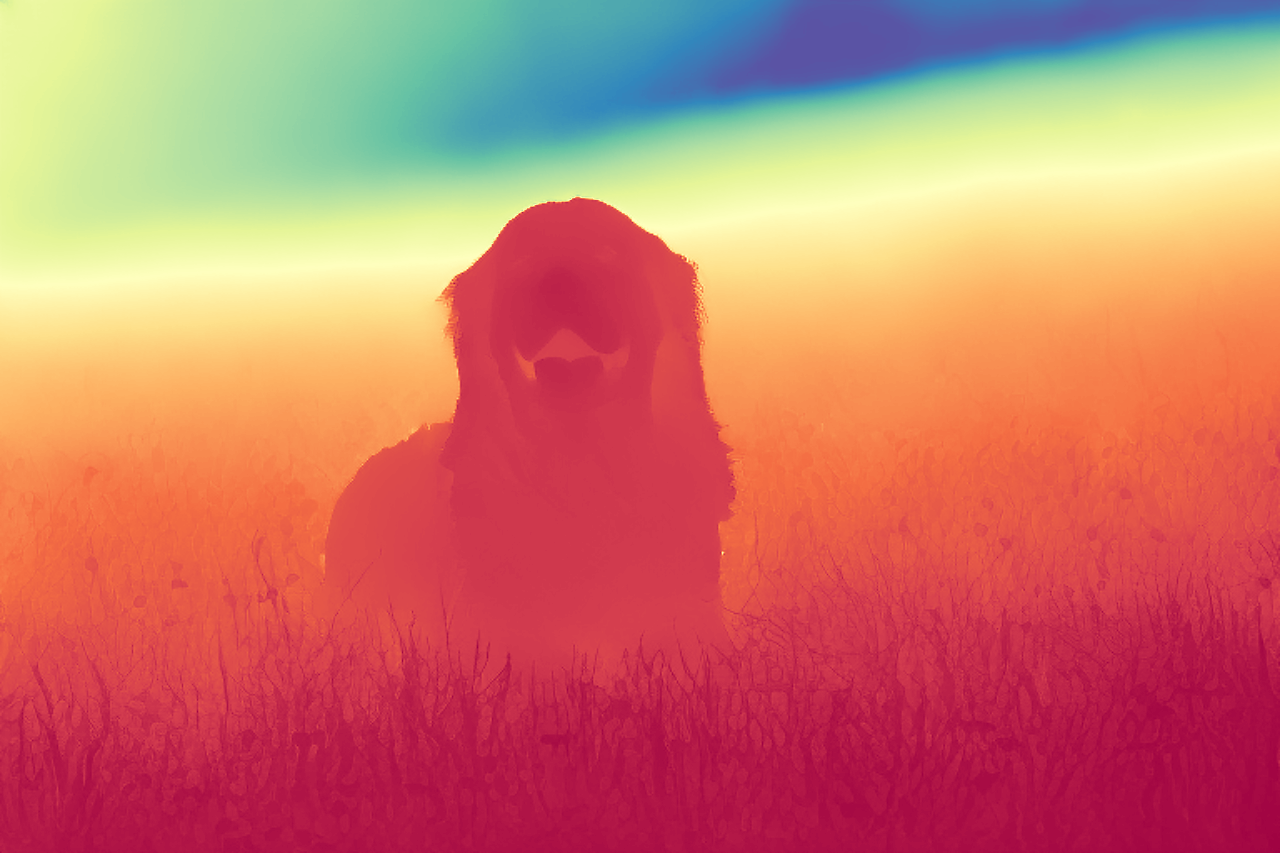}
    \end{subfigure}\hfill 
    \begin{subfigure}[]{0.185\linewidth}\centering
        \includegraphics[width=\linewidth]{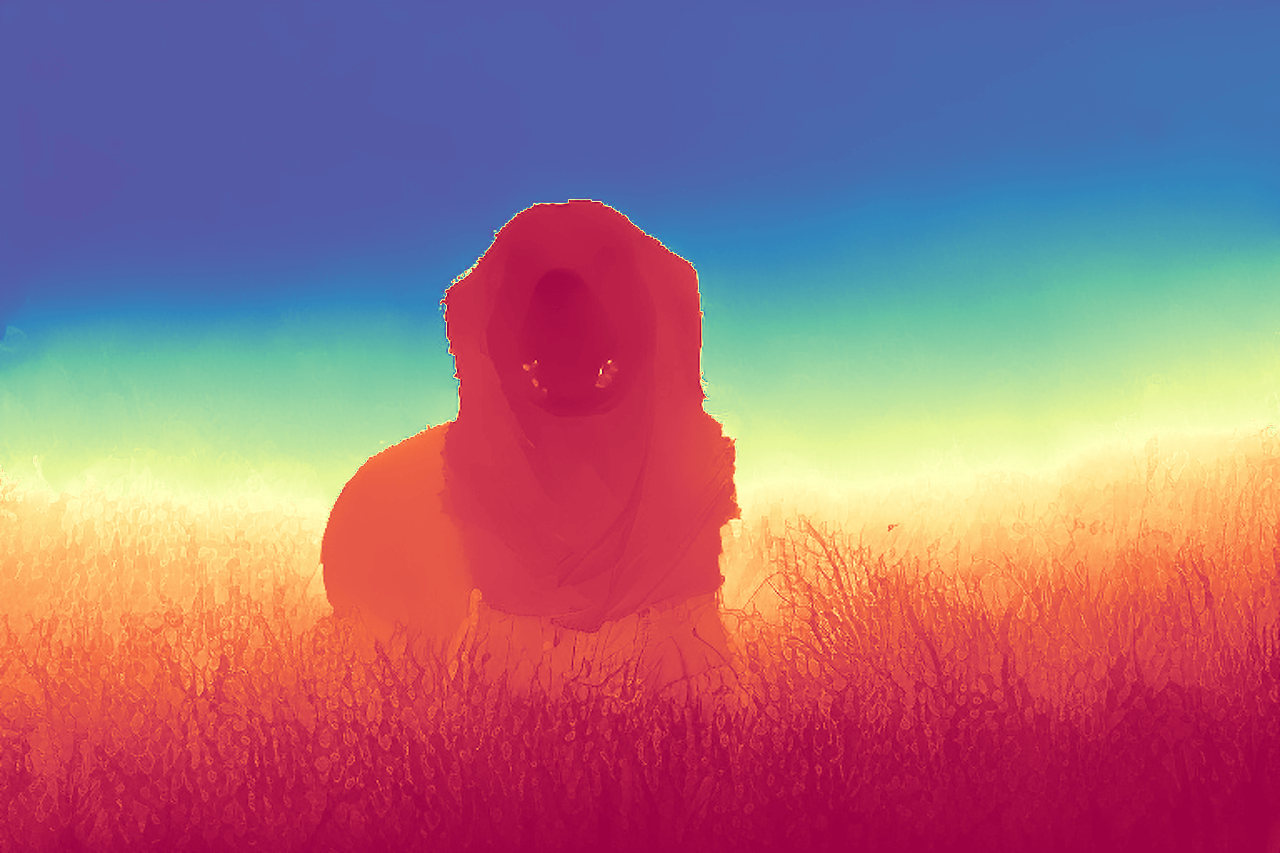}
    \end{subfigure}\hfill 
    \begin{subfigure}[]{0.185\linewidth}\centering
        \includegraphics[width=\linewidth]{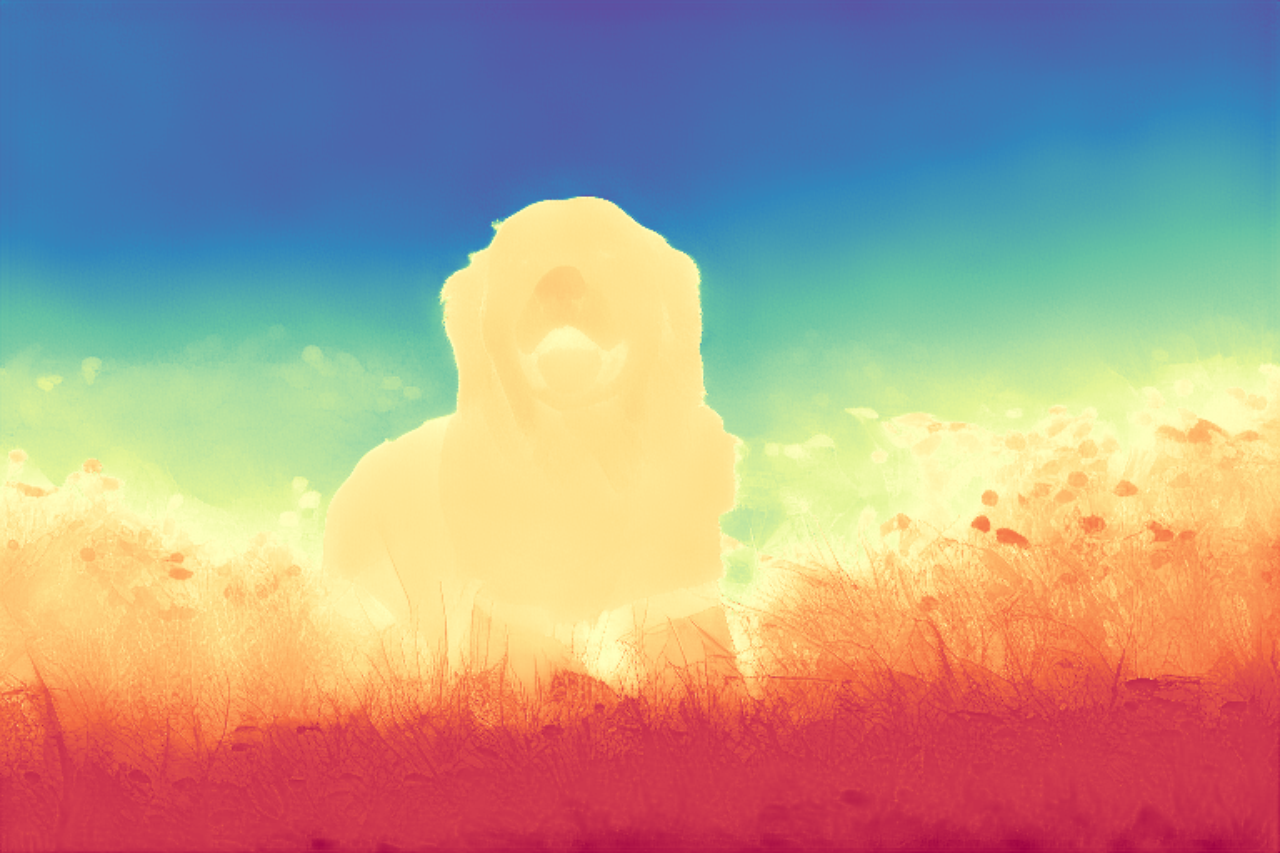}
    \end{subfigure}\vfill

\end{subfigure}
\caption{Qualitative comparison with diffusion-based depth estimation methods on in-the-wild images. DPBridge can generate depth maps with sharper boundaries and more coherent scene geometry.}
\label{fig:depth vis}
\vspace{5pt}
\begin{subfigure}{\linewidth}  \centering
    \begin{minipage}[]{0.185\linewidth}\centering \fontsize{8pt}{9.6pt}\selectfont
        Input Image        
    \end{minipage}\hfill
    \begin{minipage}[]{0.185\linewidth}\centering \fontsize{8pt}{9.6pt}\selectfont
        DPBridge
    \end{minipage}\hfill
    \begin{minipage}[]{0.185\linewidth}\centering \fontsize{8pt}{9.6pt}\selectfont
        Marigold
    \end{minipage}\hfill
    \begin{minipage}[]{0.185\linewidth}\centering \fontsize{8pt}{9.6pt}\selectfont
        Geowizard
    \end{minipage}\hfill
    \begin{minipage}[]{0.185\linewidth}\centering \fontsize{8pt}{9.6pt}\selectfont
        StableNormal
    \end{minipage}\vfill %

    \begin{subfigure}[]{0.185\linewidth}\centering
        \includegraphics[width=\linewidth]{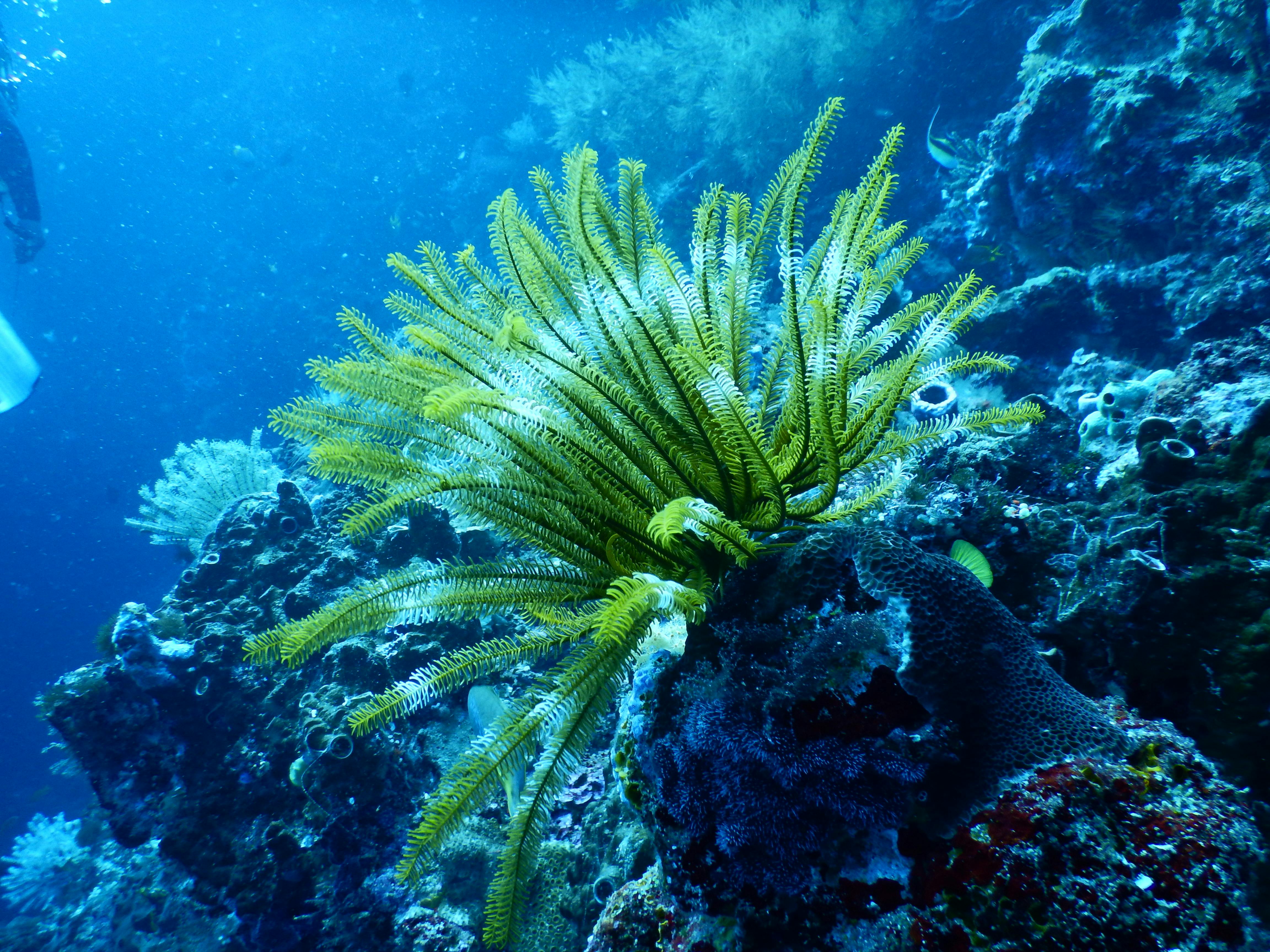}
    \end{subfigure}\hfill
    \begin{subfigure}[]{0.185\linewidth}\centering
        \includegraphics[width=\linewidth]{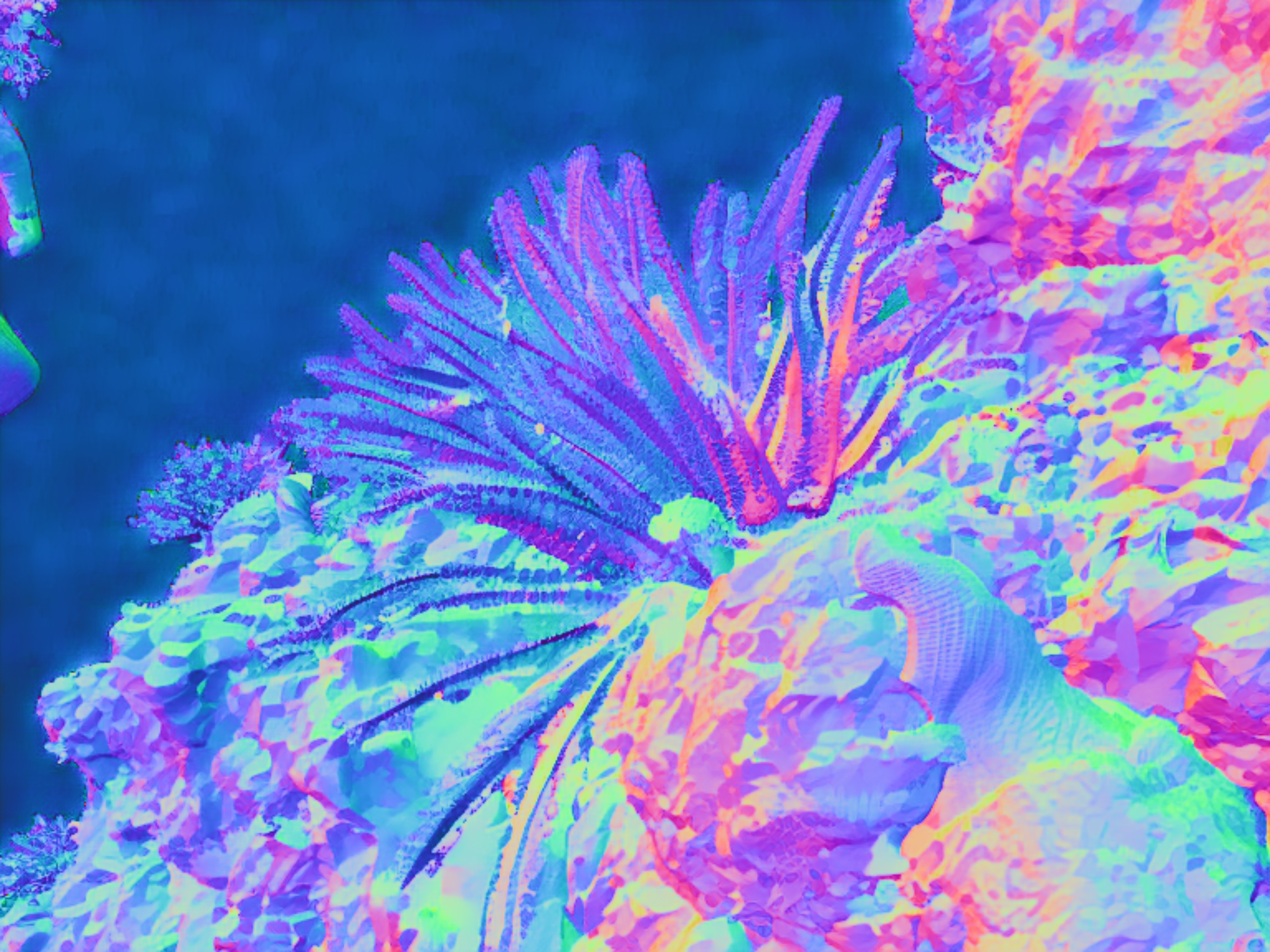}
    \end{subfigure}\hfill 
    \begin{subfigure}[]{0.185\linewidth}\centering
        \includegraphics[width=\linewidth]{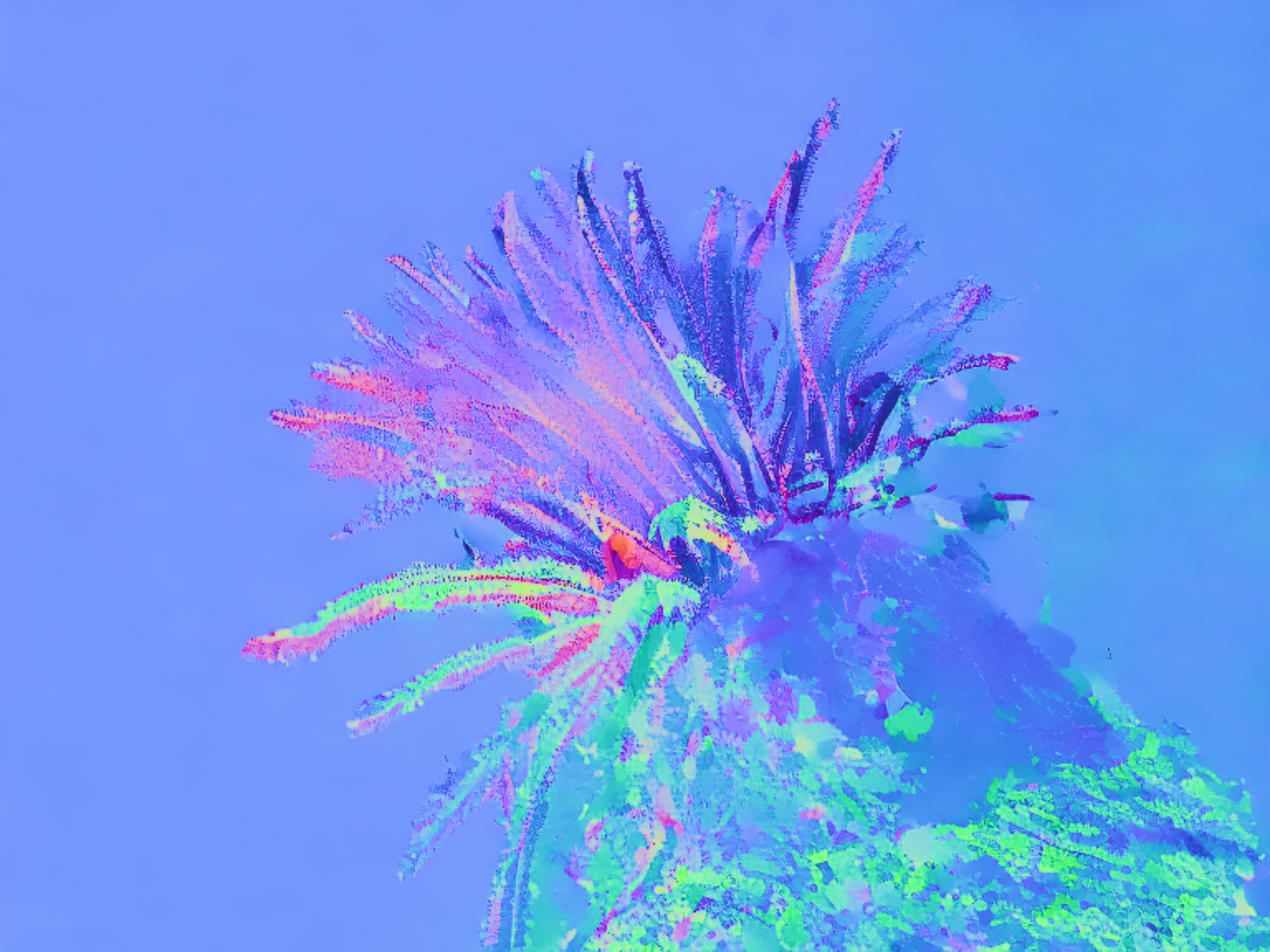}
    \end{subfigure}\hfill 
    \begin{subfigure}[]{0.185\linewidth}\centering
        \includegraphics[width=\linewidth]{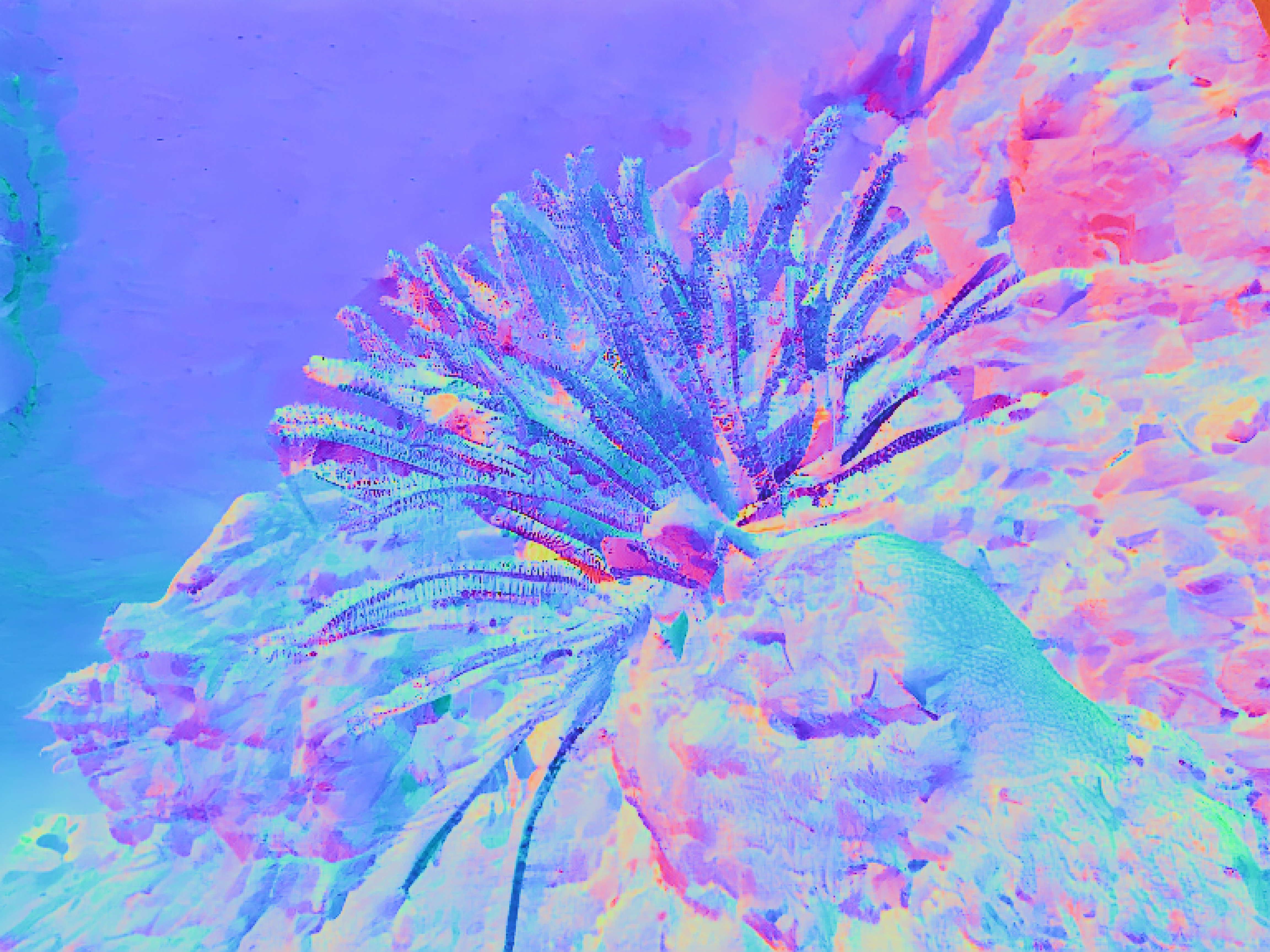}
    \end{subfigure}\hfill 
    \begin{subfigure}[]{0.185\linewidth}\centering
        \includegraphics[width=\linewidth]{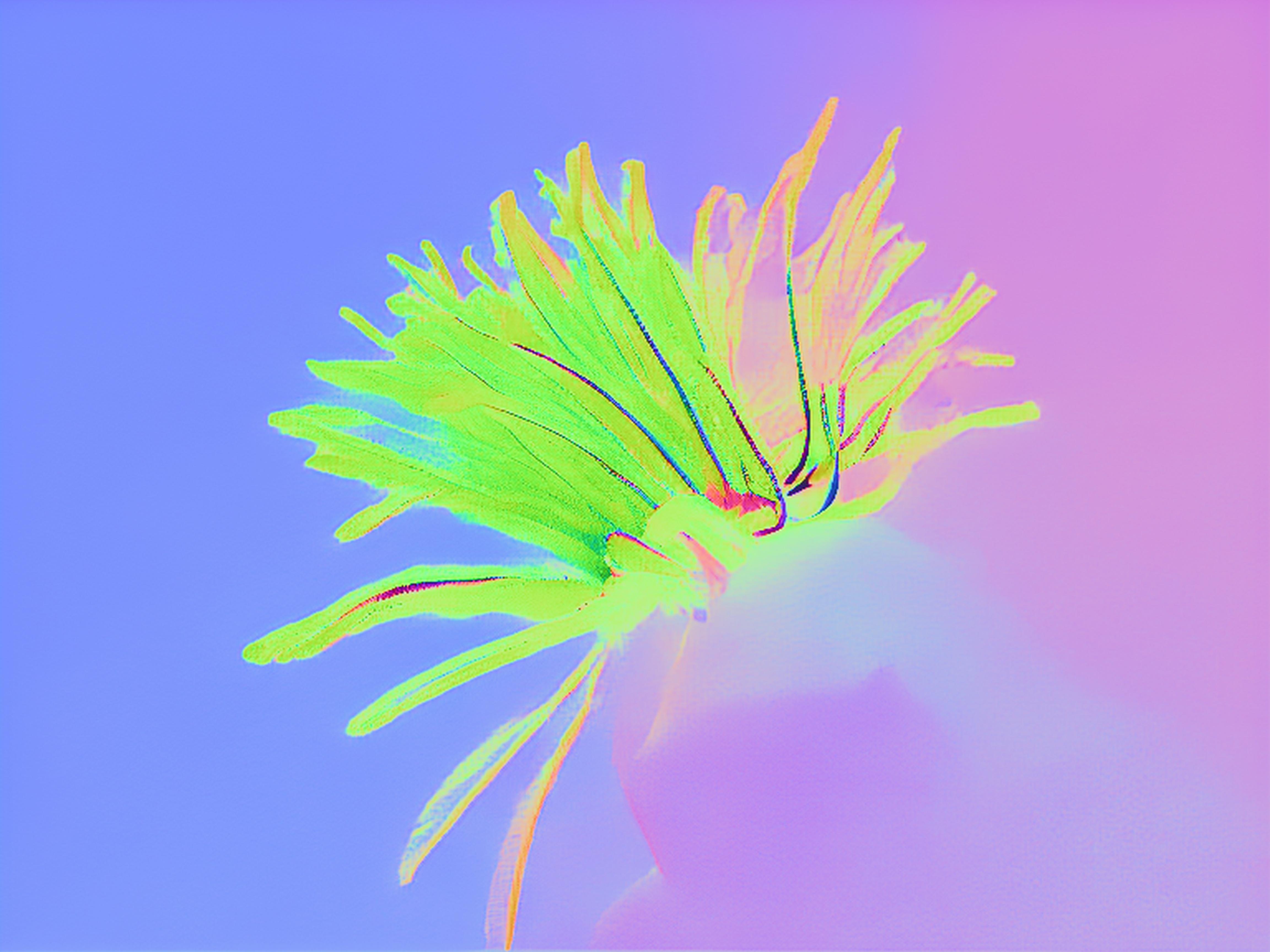}
    \end{subfigure}\vfill

    \begin{subfigure}[]{0.185\linewidth}\centering
        \includegraphics[width=\linewidth]{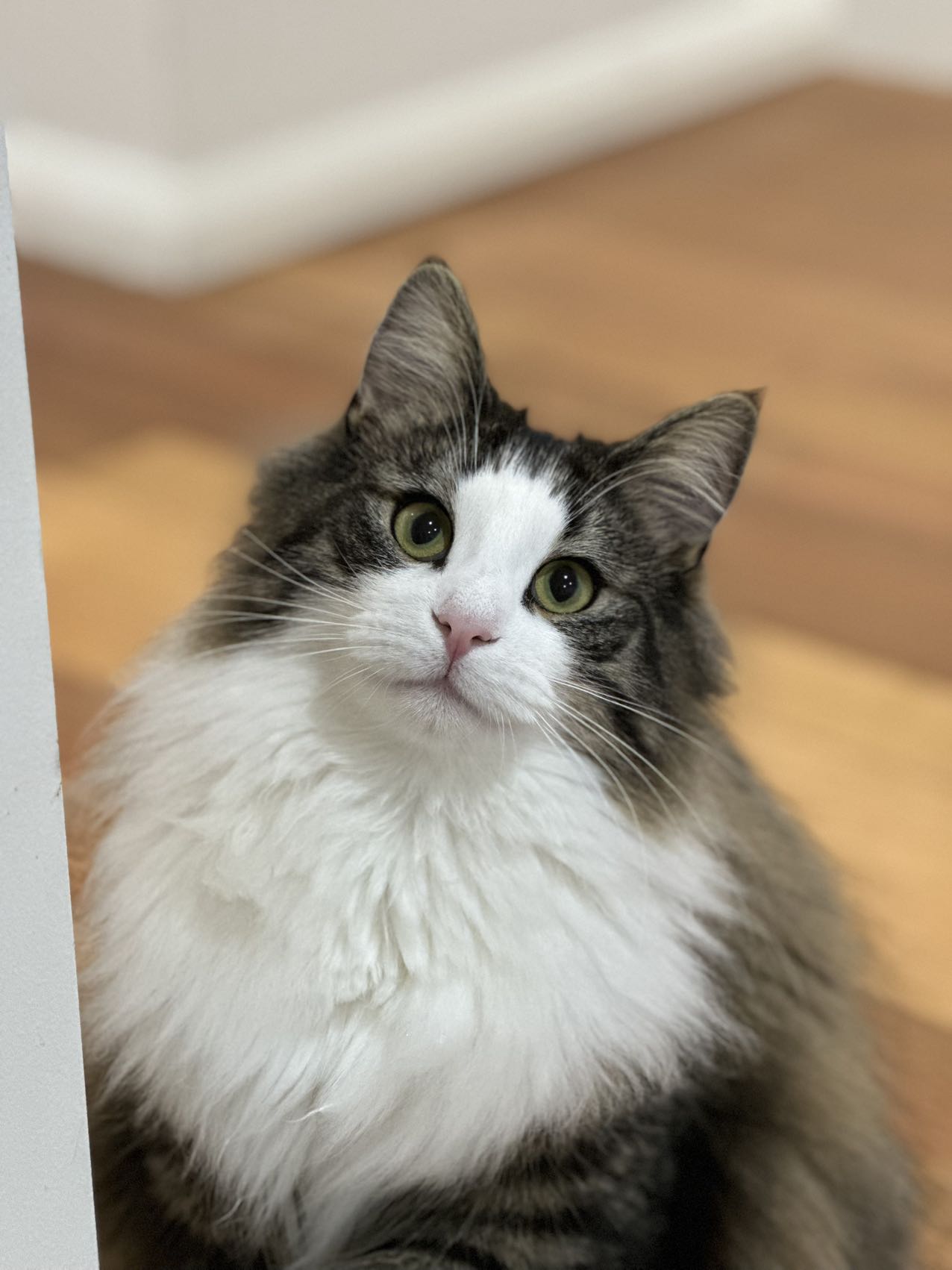}
    \end{subfigure}\hfill
    \begin{subfigure}[]{0.185\linewidth}\centering
        \includegraphics[width=\linewidth]{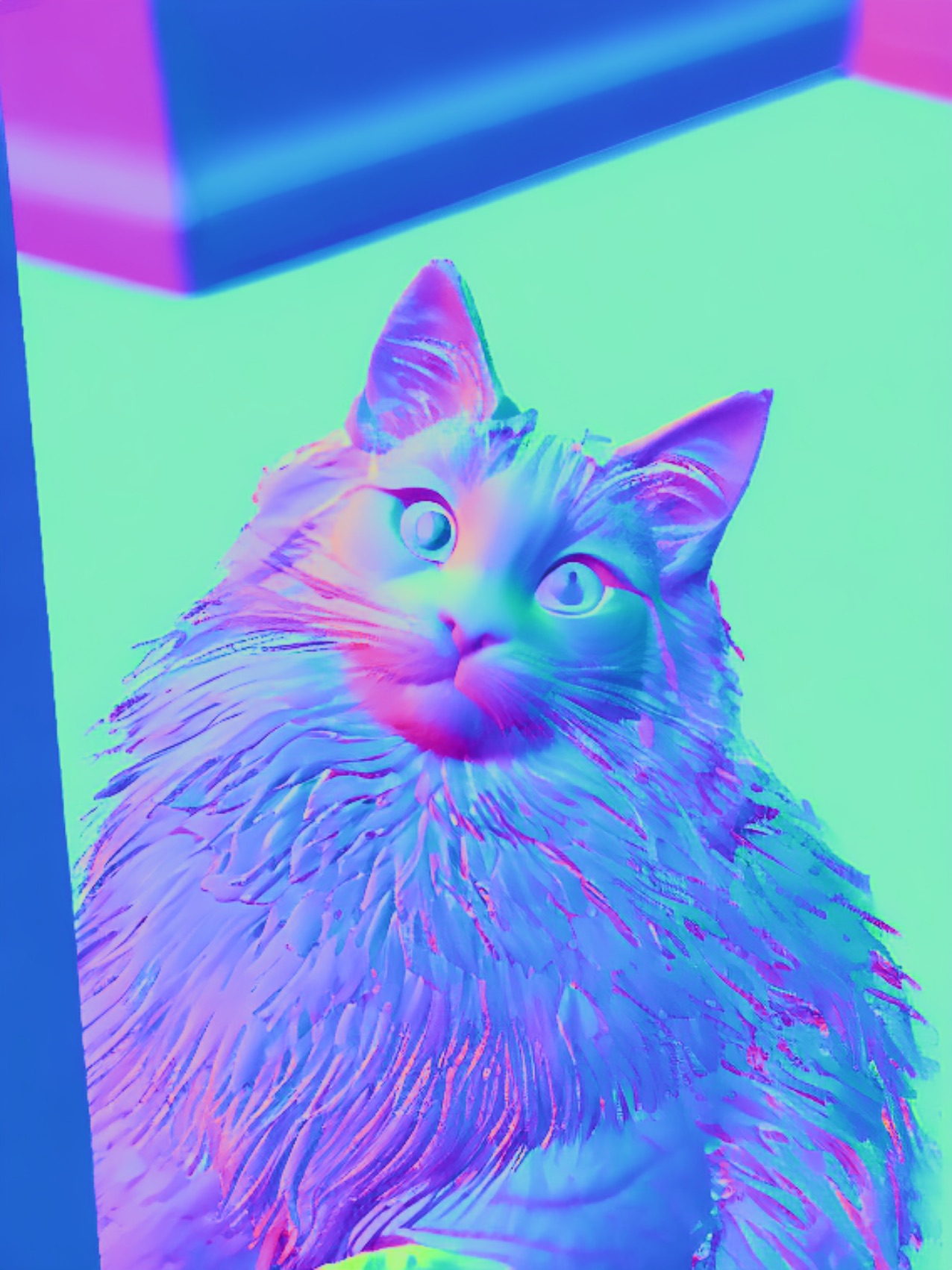}
    \end{subfigure}\hfill 
    \begin{subfigure}[]{0.185\linewidth}\centering
        \includegraphics[width=\linewidth]{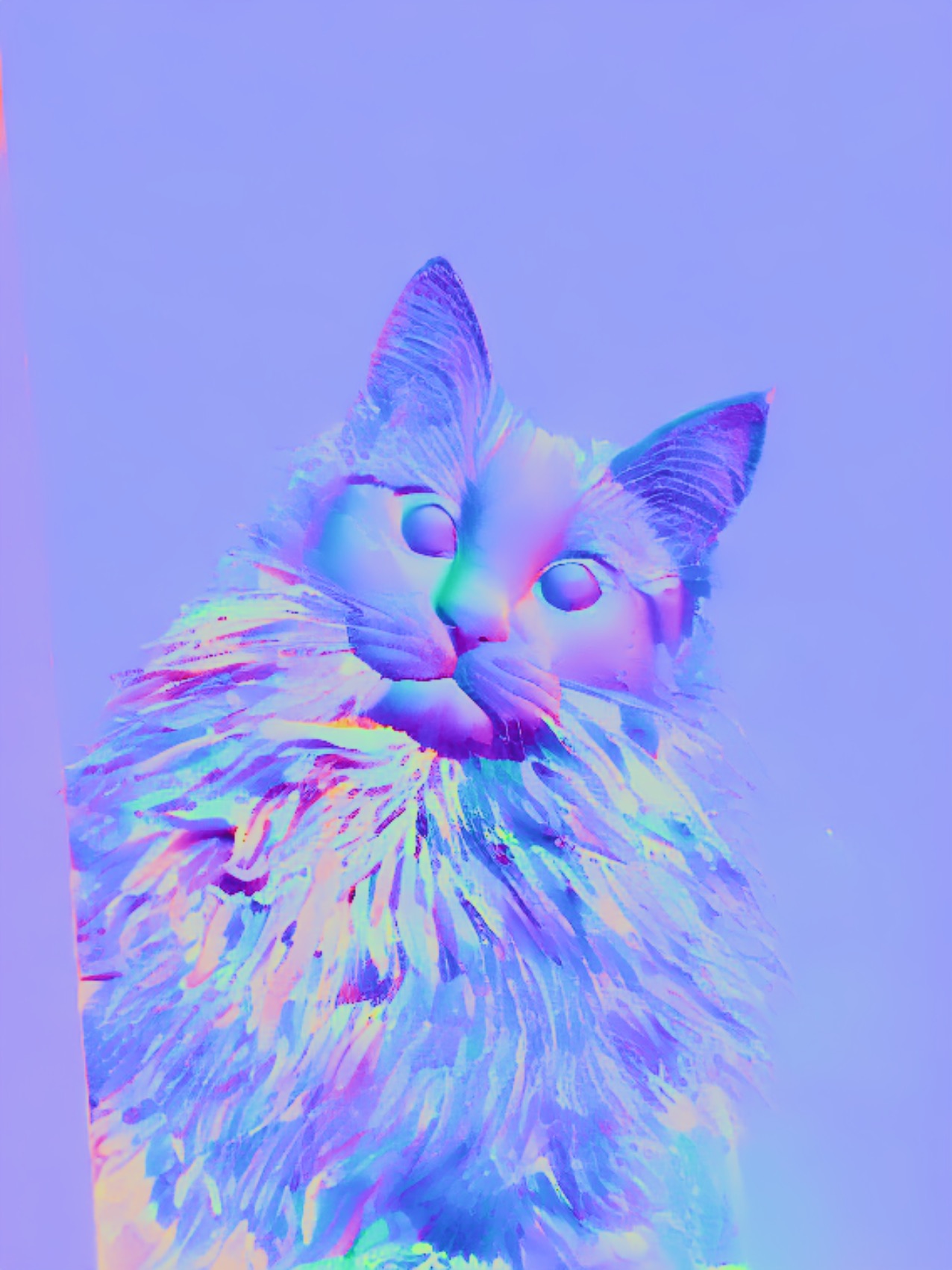}
    \end{subfigure}\hfill 
    \begin{subfigure}[]{0.185\linewidth}\centering
        \includegraphics[width=\linewidth
        ]{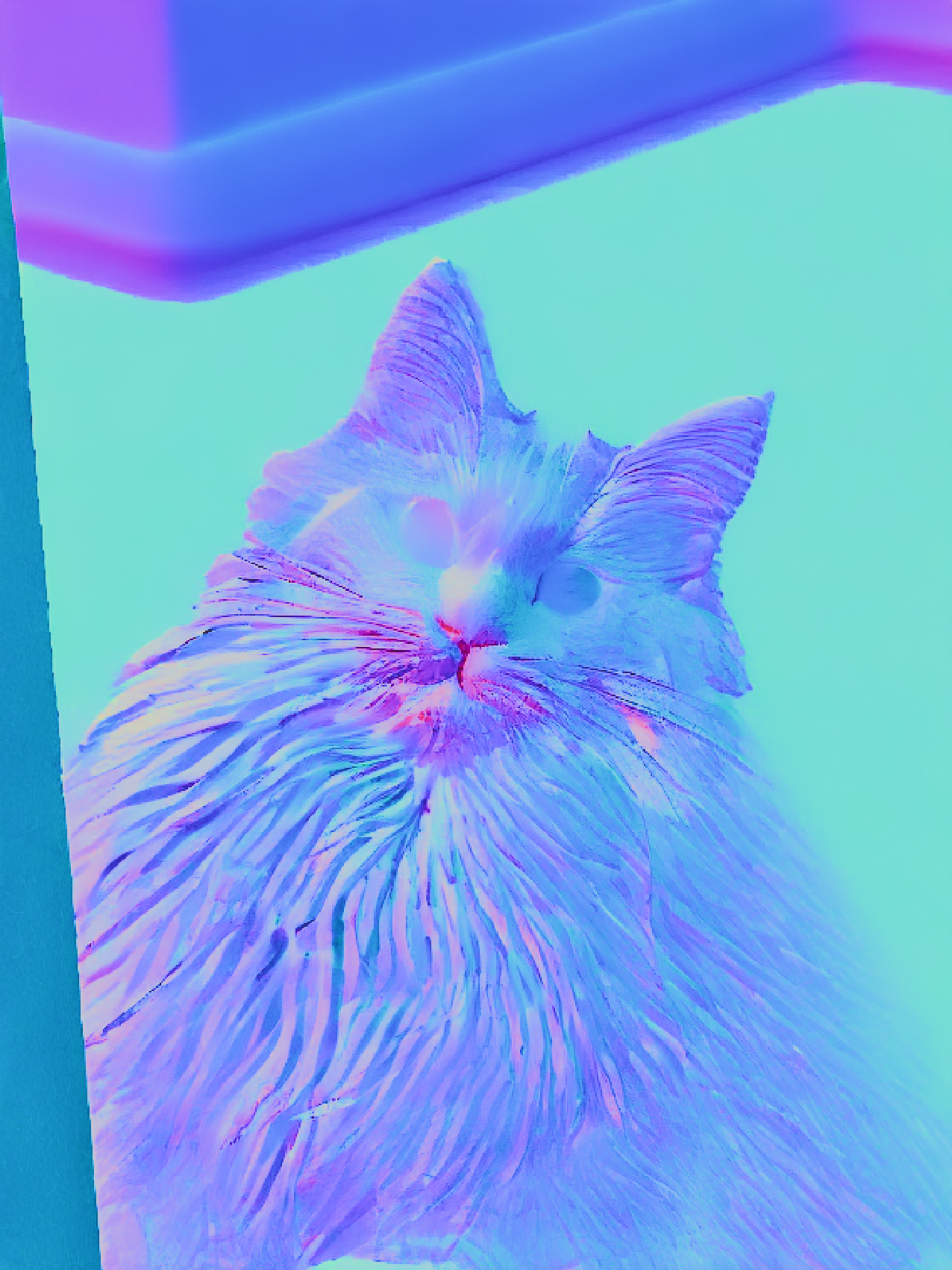}
    \end{subfigure}\hfill 
    \begin{subfigure}[]{0.185\linewidth}\centering
        \includegraphics[width=\linewidth
        ]{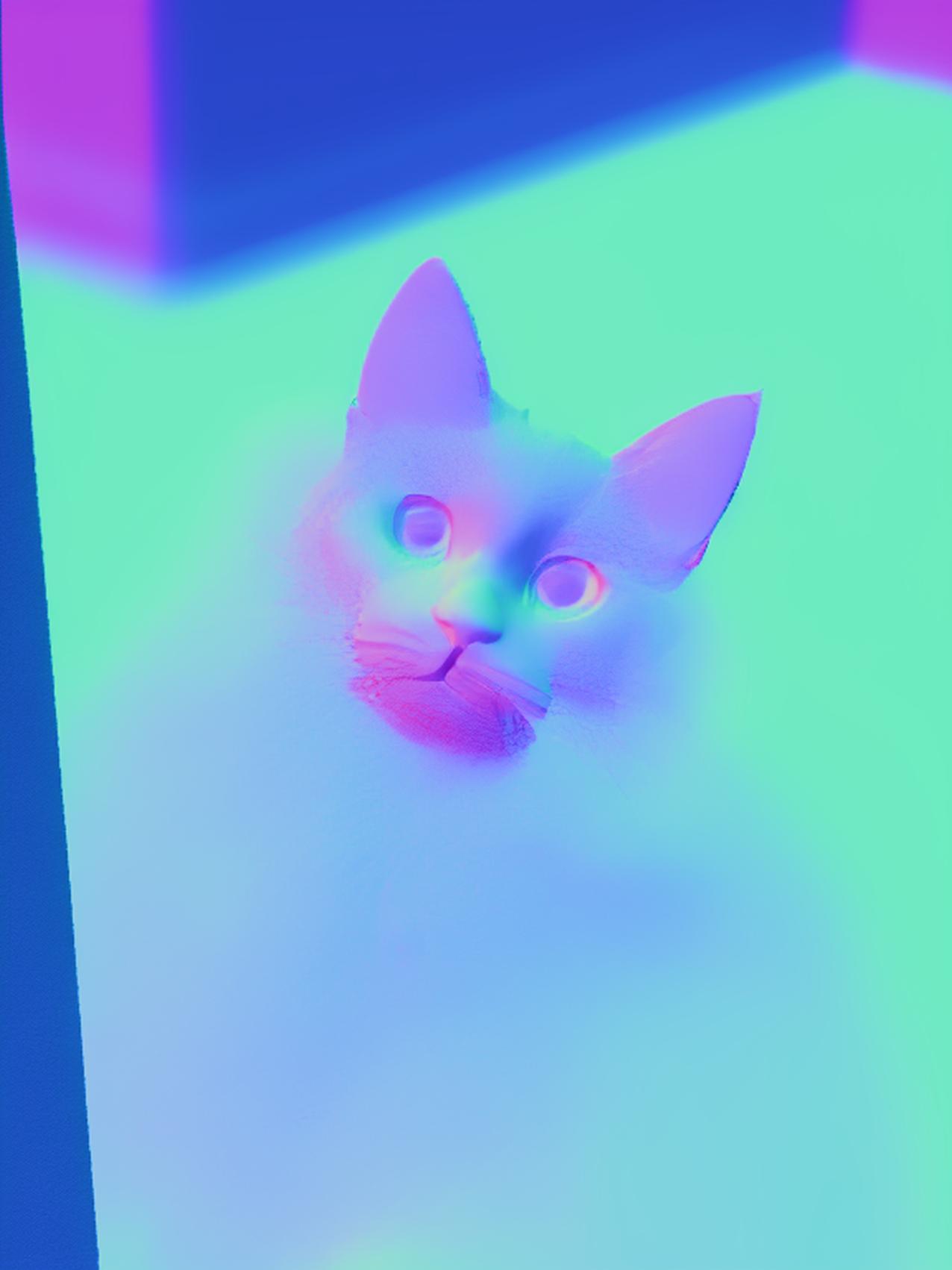}
    \end{subfigure}\vfill

    \begin{subfigure}[]{0.185\linewidth}\centering
        \includegraphics[width=\linewidth]{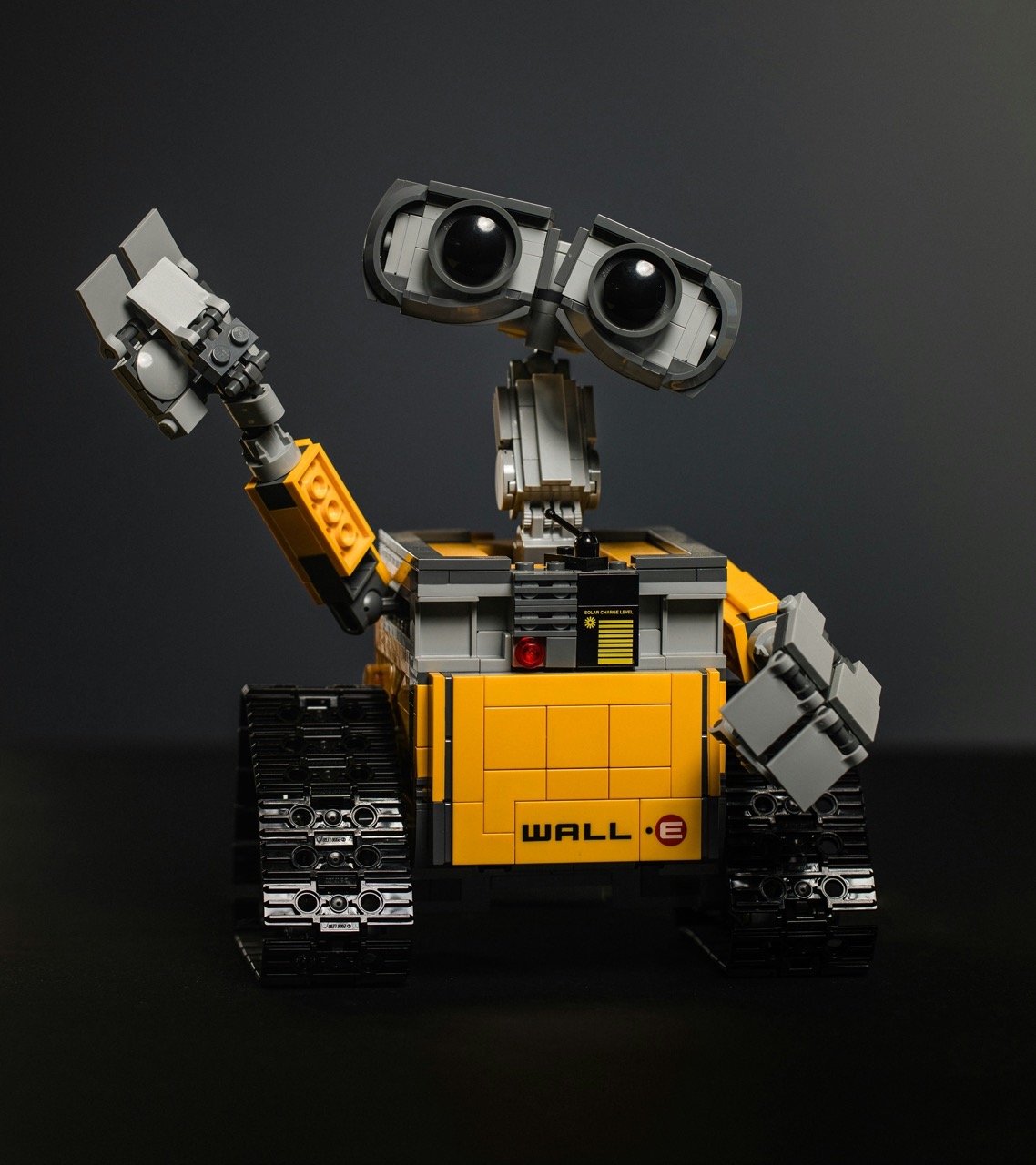}
    \end{subfigure}\hfill
    \begin{subfigure}[]{0.185\linewidth}\centering
        \includegraphics[width=\linewidth]{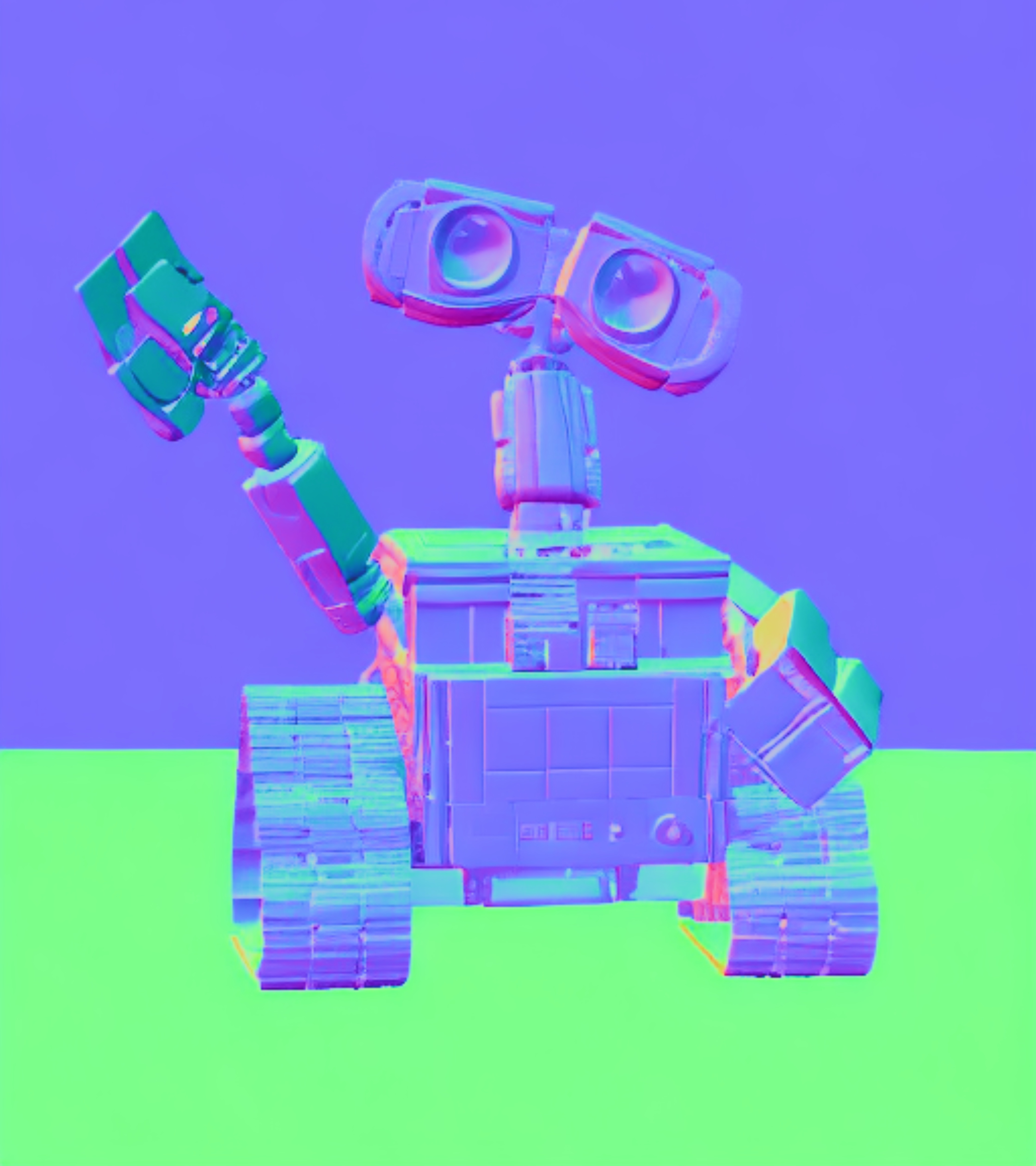}
    \end{subfigure}\hfill 
    \begin{subfigure}[]{0.185\linewidth}\centering
        \includegraphics[width=\linewidth]{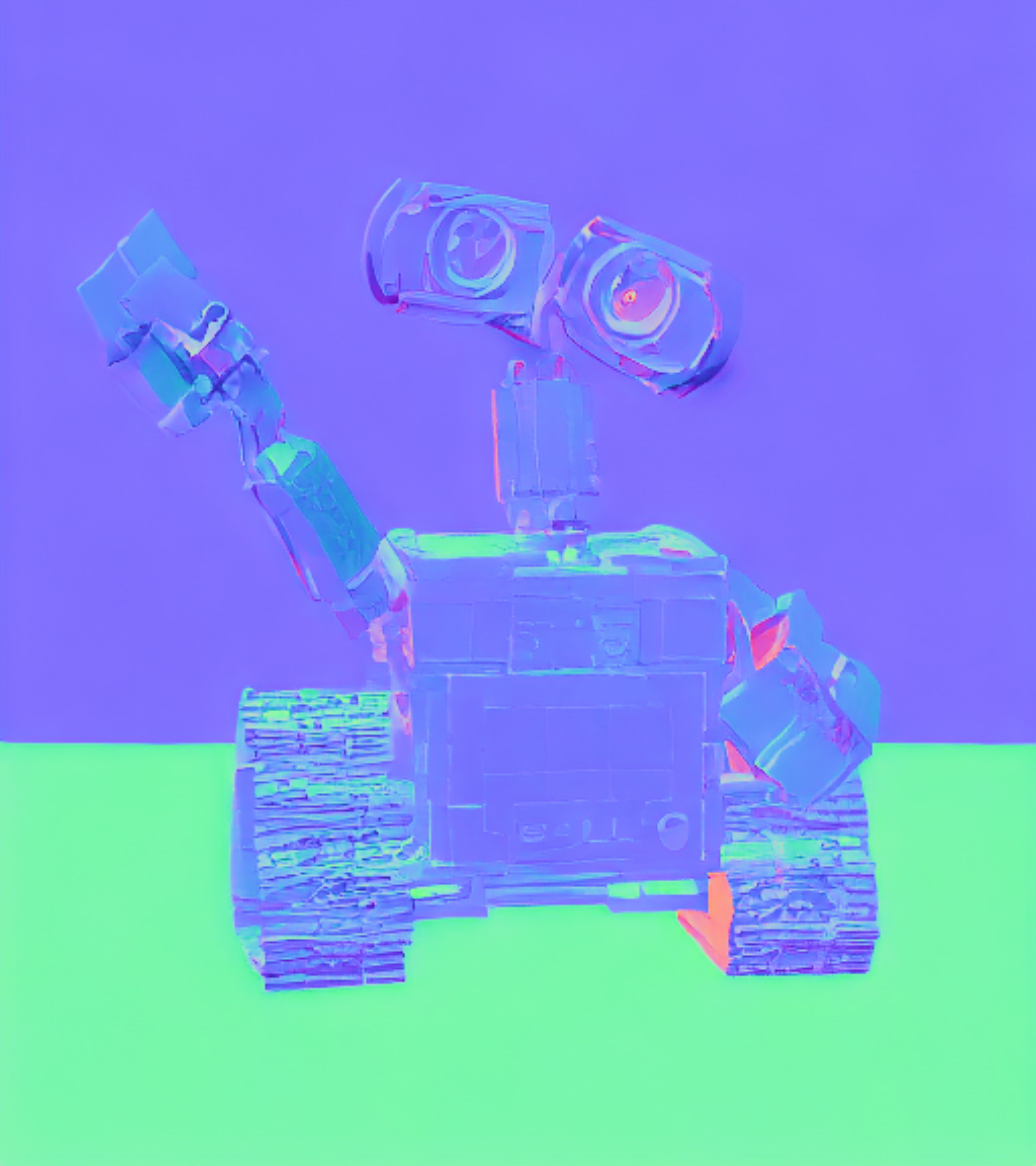}
    \end{subfigure}\hfill 
    \begin{subfigure}[]{0.185\linewidth}\centering
        \includegraphics[width=\linewidth]{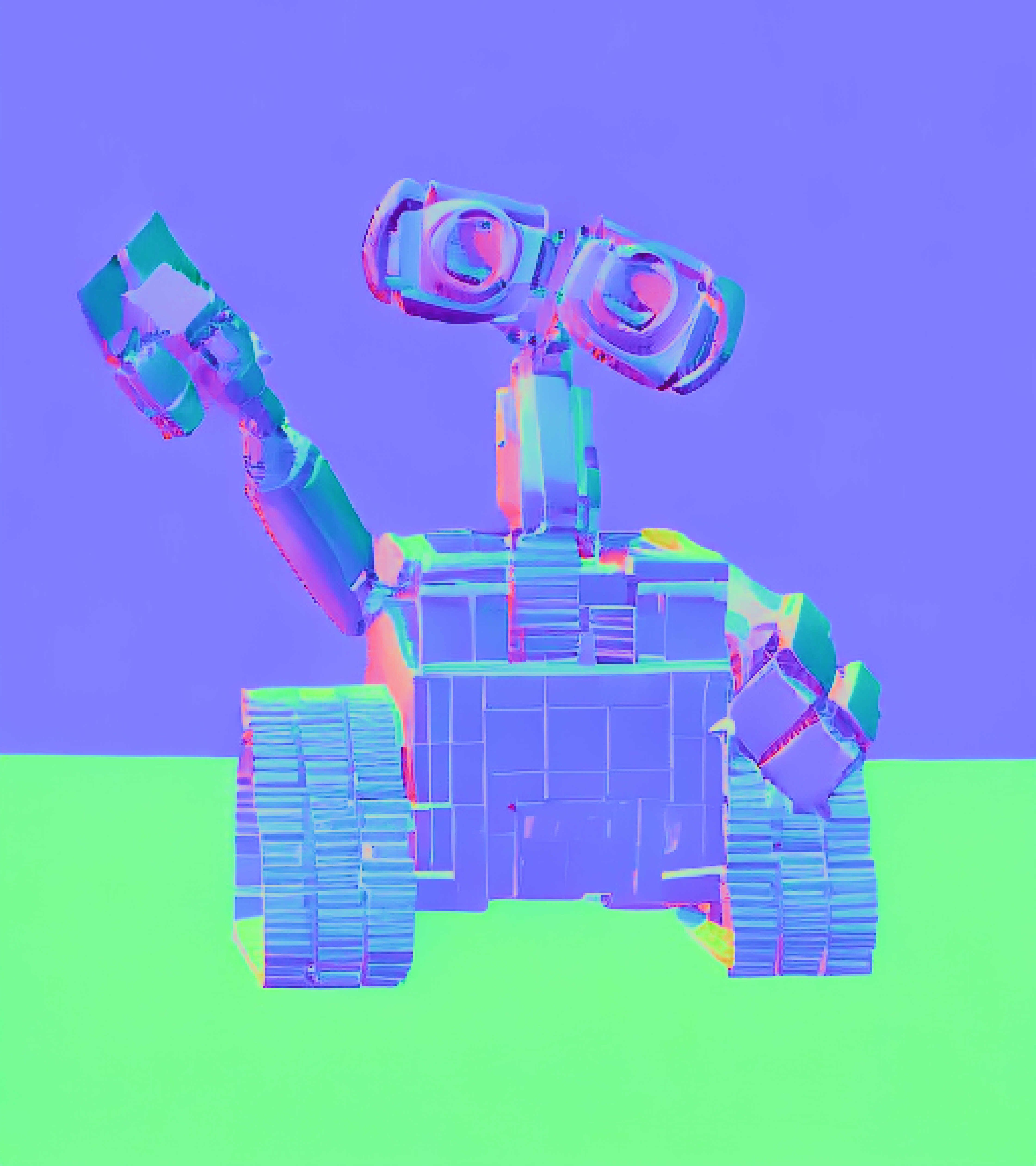}
    \end{subfigure}\hfill 
    \begin{subfigure}[]{0.185\linewidth}\centering
        \includegraphics[width=\linewidth]{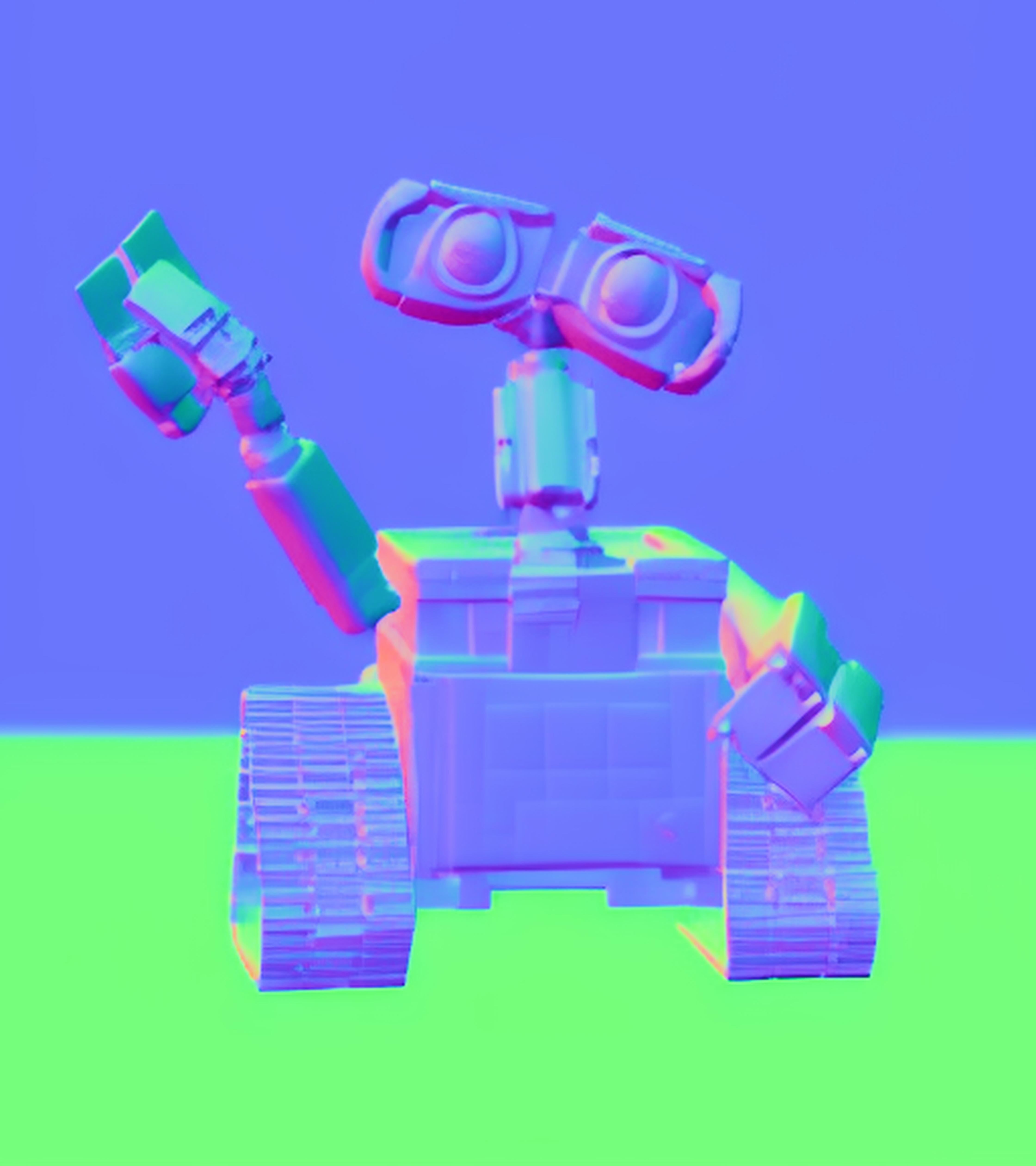}
    \end{subfigure}\vfill

\end{subfigure}
    \caption{Qualitative comparison with diffusion-based surface normal prediction methods on in-the-wild images. DPBridge exhibits more fine-grained geometric details, such as the coral and reef's contour, the kitten's fur details, and the robot's body composition curvature.}
    \label{fig:normal_map vis}
\end{figure*}

\subsection{Quantitative and Qualitative Results}
\cref{tab:depth estimation} and \cref{tab:surface normal prediction} summarize the quantitative results of DPBridge on zero-shot benchmarks across depth estimation and surface normal prediction tasks. We evaluate two variants of our model—DPBridge (SD1.5) and DPBridge (SD2.1)—built on top of Stable Diffusion v1.5 and v2.1-base~\cite{rombach2022high}, respectively. For depth estimation, DPBridge (SD2.1) achieves an AbsRel of 4.4 and $\delta_1$ accuracy of 97.6\% on NYUv2, and 4.5 AbsRel / 97.4\% $\delta_1$ on ScanNet, both among the best in class. Across all five datasets, it achieves an average ranking of 2.7—second only to Depth Anything V2 (rank 2.3), which is trained on over 1,000 times more data. For surface normal prediction, DPBridge also delivers state-of-the-art performance with a mean angular error of 14.8° and $65.3\%$ accuracy under $11.25^{\circ}$ on NYUv2, outperforming both feed-forward and diffusion-based methods. It also maintains strong results on iBims-1 ($17.1$, $68.0\%$) and Sintel ($25.0$, $46.3\%$). These results highlight DPBridge's versatility to generalize well across tasks and benchmarks. Notably, compared to prior bridge models such as I2SB, BBDM, and DDBM, DPBridge reduces average error by a large margin (e.g., 4.4 vs. 23.5 AbsRel in NYUv2 depth), underscoring the benefits and importance of incorporating pretrained visual priors to bridge construction. Furthermore, among diffusion-based approaches, DPBridge consistently achieves the best or second-best scores, validating our core hypothesis that bridge formulation provides clear advantages over standard diffusion frameworks in dense prediction tasks, and combining bridge modeling with pretrained visual prior offers a more efficient and generalizable framework.

\cref{fig:depth vis} and \cref{fig:normal_map vis} present qualitative comparisons of DPBridge with SOTA diffusion-based methods on in-the-wild images for depth estimation and surface normal prediction, respectively. As shown in \cref{fig:depth vis}, DPBridge generates depth maps with sharper object boundaries and more coherent scene geometry, particularly in challenging cases such as urban scenes or complex foreground objects like animals, grass, and flowers. In contrast, competing methods like Geowizard and DepthFM often produce over-smoothed outputs or fail to preserve depth discontinuities. In \cref{fig:normal_map vis}, DPBridge demonstrates stronger fidelity in fine-grained geometric details, such as the contour of coral and reef, the fur details of the cat, and the body composition curvature of the robot. Competing method like Marigold tends to focus on the center object in the image and ignore the environment, while StableNormal produces unreliable outputs. These visual results reinforce the quantitative findings and highlight the effectiveness of DPBridge in modeling structured output distributions with both global coherence and fine details.

\begin{figure}
    \centering
    \includegraphics[width=\linewidth]{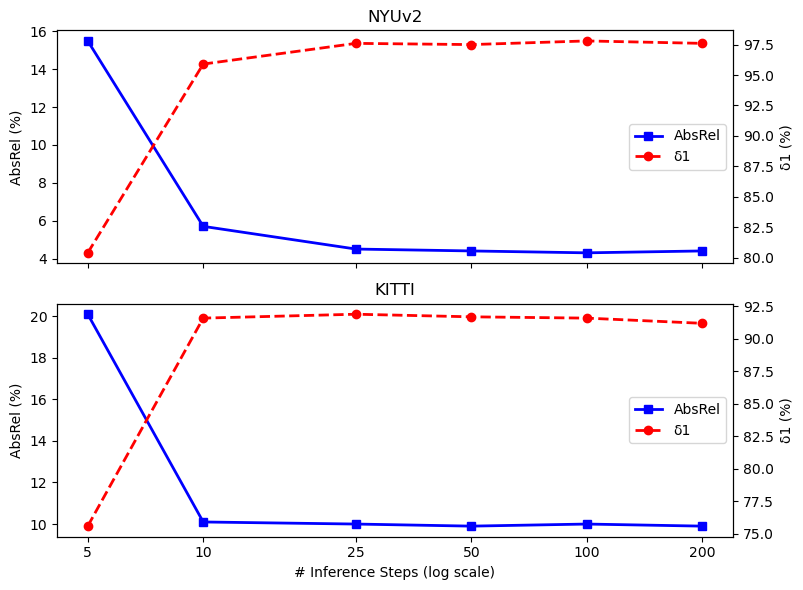}
    \caption{Ablation of sampling steps. The performance improves as the number of steps increases, while observing saturation after 10 steps.}
    \label{fig:ablation on steps}
\end{figure}

\begin{table}[t]
    \centering
    \setlength{\abovecaptionskip}{0.2cm}
    \resizebox{\linewidth}{!}{%
    \begin{tabular}{cccccccc}
    \toprule
    \multicolumn{3}{c}{\textbf{Design Components}} & \multicolumn{2}{c}{\textbf{NYUv2}} & \multicolumn{2}{c}{\textbf{KITTI}}\\ 
    \textit{FT} & \textit{DAN} & \textit{IC} & AbsRel$\downarrow$ & $\delta_1\uparrow$ & AbsRel$\downarrow$ & $\delta_1\uparrow$ \\
    \midrule
     & \checkmark & \checkmark & 21.8 & 83.7 & 30.6 & 62.7 \\
    \checkmark &  &  & 6.4 & 94.0 & 11.2 & 87.0 \\
    \checkmark &  & \checkmark & 6.2 & 95.8 & 11.0 & 87.5 \\
    \checkmark & \checkmark &  & 4.7 & 97.1 & 10.0 & 91.0 \\
    \midrule
    \checkmark & \checkmark & \checkmark & 4.4 & 97.6 & 9.8 & 91.6 \\
    \bottomrule
    \end{tabular}
    }
    \captionof{table}{Ablation studies regarding different design components. \textit{FT} stands for finetuning pretrained backbone. \textit{DAN} denotes distribution aligned normalization technique. \textit{IC} indicates image consistency loss.}
    \label{tab:ablation studies on design components}
\end{table}

\subsection{Ablation Studies}
To evaluate the effectiveness of each design component and training setting, we conduct ablation studies using depth estimation tasks on the NYUv2 and KITTI datasets. The results are summarized in \cref{tab:ablation studies on design components} and \cref{fig:ablation on steps}, with detailed discussions provided below.

\noindent\textbf{Visual Prior Exploitation.} To evaluate the usage of visual priors, we compare finetuning with training from scratch. As shown in the first row of \cref{tab:ablation studies on design components}, finetuning yields significantly better results under identical settings. This is attributed to stronger visual priors from foundation models and faster convergence. Additionally, the last rows of \cref{tab:depth estimation} and \cref{tab:surface normal prediction} show that SD2.1 outperforms SD1.5, highlighting that higher-quality priors further boost performance. These results confirm the advantage of building DPBridge on strong pretrained backbones.

\noindent\textbf{Distribution Alignment Normalization.} We next assess the benefit of aligning marginal distributions between the diffusion bridge and standard diffusion processes. As shown in the second and fourth rows of \cref{tab:ablation studies on design components}, adding a distribution alignment normalization step before the noise prediction network consistently improves performance.

\noindent\textbf{Image Consistency Loss.} As shown in the third and fourth rows of \cref{tab:ablation studies on design components}, the auxiliary image consistency loss improves performance but has a smaller impact than distribution alignment. This is likely because it mainly affects the boundaries or overlapping regions between masked and valid areas, which are limited in scope. Nonetheless, when combined with other components, it contributes to a notable overall performance gain.

\noindent\textbf{Number of Sampling Steps.} We evaluate performance under varying sampling steps during inference. As shown in \cref{fig:ablation on steps}, perception quality improves rapidly with more steps and saturates after 10. Notably, compared to \cref{tab:depth estimation}, where other diffusion-based methods use 50 steps, DPBridge achieves comparable results with significantly fewer steps, highlighting its efficiency.

\vspace{-0.5em}
\section{Conclusion}
\vspace{-0.5em}
\label{sec:conclusion}
In this paper, we propose DPBridge, a latent diffusion bridge framework for dense prediction tasks through integrating the strengths of diffusion bridge modeling with visual priors from pretrained foundation models. We derive a fully tractable bridge process to establish direct probability transport in the latent space, and propose finetuning strategies to better adapt pretrained diffusion backbones in bridge construction. While DPBridge achieves competitive performance across different tasks and benchmarks, several limitations remain. First, our method still requires iterative sampling, which makes it slower than feed-forward approaches. Exploring acceleration strategies such as distillation or single-step bridge solvers could further improve practical deployment, especially in real-time settings. Second, our formulation is restricted to certain bridge parameterization, and exploring alternatives to adapt to more advanced pretrained backbone is a promising future direction.

\section*{Acknowledgement}
\label{sec:acknowledgement}
This research is funded in part by an ARC (Australian Research Council) Discovery Grant of DP220100800. Prof Hongdong Li holds concurrent appointments as a Full Professor with the ANU and as an Amazon Scholar with Amazon (parttime). This paper describes work performed at ANU and is not associated with Amazon.

{
    \small
    \bibliographystyle{ieeenat_fullname}
    \bibliography{main}
}

\newpage
\appendix
\onecolumn


\section{Derivation Details}
\label{app:derivation details}
In this section, we'll provide derivation details of the maximum likelihood loss presented in \cref{sec:train loss}. Its core is the posterior distribution $q(\mathbf{z}_{t-1} | \mathbf{z}_t, \mathbf{z}_0, \mathbf{z}_T)$ as in \cref{eqn: bridge reverse kernel}, whose calculation involves three terms: $q(\mathbf{z}_t | \mathbf{z}_{t-1}, \mathbf{z}_T)$, $q(\mathbf{z}_{t-1} | \mathbf{z}_0, \mathbf{z}_T)$, and $q(\mathbf{z}_{t} | \mathbf{z}_0, \mathbf{z}_T)$. The last two terms of the posterior can be retrieved directly through the marginal distribution of the diffusion bridge's forward process, but further derivation is required for the first term. Specifically, as specified in \cref{eqn: bridge forward kernel}:
\begin{align}
\label{eqn:z_t}
\begin{split}
    q(\mathbf{z}_t | \mathbf{z}_0, \mathbf{z}_T) &= \mathcal{N}(m_t\mathbf{z}_0 + n_t\mathbf{z}_T, \sigma_t^2 \mathbf{I})  \\
    \mathbf{z}_t &= m_t\mathbf{z}_0 + n_t\mathbf{z}_T + \sigma_t\boldsymbol{\epsilon}_t 
\end{split}
\\[2ex]
\label{eqn:z_t-1}
\begin{split}
    q(\mathbf{z}_{t-1} | \mathbf{z}_0, \mathbf{z}_T) &= \mathcal{N}(m_{t-1}\mathbf{z}_0 + n_{t-1}\mathbf{z}_T, \sigma_{t-1}^2 \mathbf{I})  \\
    \mathbf{z}_{t-1} &= m_{t-1}\mathbf{z}_0 + n_{t-1}\mathbf{z}_T + \sigma_{t-1}\boldsymbol{\epsilon}_{t-1}
\end{split}
\end{align}
To derive the transition probability $q(\mathbf{z}_t | \mathbf{z}_{t-1}, \mathbf{z}_T)$, we substitute $\mathbf{z}_0$ in \cref{eqn:z_t} with derivation from \cref{eqn:z_t-1}, obtain
\begin{equation}
\begin{aligned}
    \mathbf{z}_t &= a_t\mathbf{z}_{t-1} + b_t\mathbf{z}_T + \delta_t\boldsymbol{\epsilon}  \\
    a_t &= \frac{m_t}{m_{t-1}} \\
    b_t &= n_t - \frac{m_t}{m_{t-1}} n_{t-1} = n_t - a_t n_{t-1} \\
    \delta_t^2 &= \sigma_t^2 - \frac{m_t}{m_{t-1}} n_{t-1}^2\sigma_{t-1}^2 = \sigma_t^2 - a_t^2\sigma_{t-1}^2
\end{aligned}
\end{equation}
Thus, we can get $q(\mathbf{z}_t | \mathbf{z}_{t-1}, \mathbf{z}_T) = \mathcal{N}(a_t\mathbf{z}_{t-1} + b_t\mathbf{z}_T, \delta_t^2 \mathbf{I})$, which is also a Gaussian distribution.

\noindent Putting all these terms back to \cref{eqn: bridge reverse kernel}, we'll have:
\begin{equation}
\begin{aligned}
    &q(\mathbf{z}_{t-1} | \mathbf{z}_t, \mathbf{z}_0, \mathbf{z}_T) \sim \mathcal{N}(\tilde{\boldsymbol{\mu}}_t(\mathbf{z}_t, \mathbf{z}_0, \mathbf{z}_T), \tilde{\sigma}_t^2\mathbf{I})  \\
  = &\frac{q(\mathbf{z}_t | \mathbf{z}_{t-1}, \mathbf{z}_T)q(\mathbf{z}_{t-1} | \mathbf{z}_0, \mathbf{z}_T)}{q(\mathbf{z}_t | \mathbf{z}_0, \mathbf{z}_T)} \\
  \propto &\exp \bigg[
        -\frac{1}{2} \bigg(
            \frac{(\mathbf{z}_t - a_t\mathbf{z}_{t-1} - b_t\mathbf{z}_T)^2}{\delta_t^2} 
            + \frac{(\mathbf{z}_{t-1} - m_{t-1}\mathbf{z}_0 - n_{t-1}\mathbf{z}_T)^2}{\sigma_{t-1}^2} 
            - \frac{(\mathbf{z}_t - m_t\mathbf{z}_0 - n_t\mathbf{z}_T)^2}{\sigma_t^2} 
        \bigg)
    \bigg] \\
  = &\exp \bigg[
        -\frac{1}{2} \bigg(
            \bigg(\frac{a_t^2}{\delta_t^2} + \frac{1}{\sigma_{t-1}^2}\bigg) \mathbf{z}_{t-1}^2 
            - 2 \bigg(\frac{a_t\mathbf{z}_t - a_tb_t\mathbf{z}_T}{\delta_t^2} + \frac{m_{t-1}\mathbf{z}_0 + n_{t-1}\mathbf{z}_T}{\sigma_{t-1}^2}\bigg) \mathbf{z}_{t-1} 
            + C(\mathbf{z}_t, \mathbf{z}_0, \mathbf{z}_T)
        \bigg)
    \bigg].
\end{aligned}
\end{equation}
where $C(\mathbf{z}_t, \mathbf{z}_0, \mathbf{z}_T)$ indicates the constant term that is independent of $\mathbf{z}_{t-1}$. Since the Gaussian distribution probability density $\mathbf{z}_{t-1} \sim \mathcal{N}(\tilde{\boldsymbol{\mu}}_t, \tilde{\sigma}_t^2\mathbf{I})$ can be expressed in the following form:
\begin{equation}
\begin{aligned}
    \exp\left(-\frac{(\mathbf{z}_{t-1} - \tilde{\boldsymbol{\mu}}_t)^2}{2 \tilde{\sigma}_t^2}\right)
    = \exp\left[
    -\frac{1}{2} \left(
        \frac{1}{\tilde{\sigma}_t^2}\mathbf{z}_{t-1}^2 - \frac{2\tilde{\boldsymbol{\mu}}_t}{\tilde{\sigma}_t^2}\mathbf{z}_{t-1} + \frac{\tilde{\boldsymbol{\mu}}_t^2}{\tilde{\sigma}_t^2}
    \right)\right].
\end{aligned}
\end{equation}
We can thus derive: 
\begin{align}
\label{eqn:sigma_t'}
    \frac{1}{\tilde{\sigma}_t^2} &= \frac{a_t^2}{\delta_t^2} + \frac{1}{\sigma_{t-1}^2} \nonumber\\
    \Rightarrow \tilde{\sigma}_t^2 &= \frac{\sigma_{t-1}^2(\sigma_t^2 - a_t^2\sigma_{t-1}^2)}{\sigma_t^2} = \frac{\sigma_{t-1}^2}{\sigma_t^2}\delta_t^2  \\
\frac{\tilde{\boldsymbol{\mu}}_t}{\tilde{\sigma}_t^2} &= \frac{a_t\mathbf{z}_t - a_tb_t\mathbf{z}_T}{\delta_t^2} + \frac{m_{t-1}\mathbf{z}_0 + n_{t-1}\mathbf{z}_T}{\sigma_{t-1}^2} \nonumber\\
\Rightarrow \tilde{\boldsymbol{\mu}}_t &= \bigg[\frac{\sigma_{t-1}^2\big(a_t\mathbf{z}_t - a_tb_t\mathbf{z}_T\big) + \delta_t^2\big(m_{t-1}\mathbf{z}_0 + n_{t-1}\mathbf{z}_T\big)}{\delta_t^2\sigma_{t-1}^2}\bigg]\frac{\sigma_{t-1}^2}{\sigma_t^2}\delta_t^2 \nonumber\\
&= \frac{\sigma_{t-1}^2}{\sigma_t^2} a_t \mathbf{z}_t + \frac{\delta_t^2}{\sigma_t^2} m_{t-1} \mathbf{z}_0 + \bigg(\frac{\delta_t^2}{\sigma_t^2} n_{t-1} - \frac{\sigma_{t-1}^2}{\sigma_t^2} a_t b_t\bigg) \mathbf{z}_T \label{eqn:mu_t'}
\end{align}
whose results are equivalent to \cref{eqn: bridge reverse mean} and \cref{eqn: bridge reverse variance}.

Accordingly, maximizing the Evidence Lower Bound (ELBO) is equivalent to minimizing the KL divergence between ground truth and predicted posterior:
\begin{equation}
\begin{aligned}
    &KL \left( q(\mathbf{z}_{t-1} | \mathbf{z}_0, \mathbf{z}_t, \mathbf{z}_T) \,||\, p_\theta(\mathbf{z}_{t-1} \mid \mathbf{z}_t, \mathbf{z}_T) \right) \\
= &\mathbb{E}_{q(\mathbf{z}_{t-1} | \mathbf{z}_0, \mathbf{z}_t, \mathbf{z}_T)} \left[
    \log \frac{
        \frac{1}{\sqrt{2\pi}\tilde{\sigma}_{t-1}} e^{-\frac{(\mathbf{z}_{t-1} - \tilde{\boldsymbol{\mu}}_{t-1})^2}{2\tilde{\sigma}_{t-1}^2}}
    }{
        \frac{1}{\sqrt{2\pi}\tilde{\sigma}_{\theta, t-1}} e^{-\frac{(\mathbf{z}_{t-1} - \tilde{\boldsymbol{\mu}}_{\theta, t-1})^2}{2\tilde{\sigma}_{\theta, t-1}^2}}
    }
\right] \\
= &\mathbb{E}_{q(\mathbf{z}_{t-1} | \mathbf{z}_0, \mathbf{z}_t, \mathbf{z}_T)} \left[
    \log \tilde{\sigma}_{\theta, t-1} - \log \tilde{\sigma}_{t-1} 
    - \frac{(x_{t-1} - \tilde{\boldsymbol{\mu}}_{t-1})^2}{2\tilde{\sigma}_{t-1}^2} 
    + \frac{(x_{t-1} - \tilde{\boldsymbol{\mu}}_{\theta, t-1})^2}{2\tilde{\sigma}_{\theta, t-1}^2}
\right] \\
= &\log \frac{\tilde{\sigma}_{\theta, t-1}}{\tilde{\sigma}_{t-1}} - \frac{1}{2} 
+ \frac{\tilde{\sigma}_{t-1}^2}{2\tilde{\sigma}_{\theta, t-1}^2} 
+ \frac{(\tilde{\boldsymbol{\mu}}_{t-1} - \tilde{\boldsymbol{\mu}}_{\theta, t-1})^2}{2\tilde{\sigma}_{\theta, t-1}^2}
\end{aligned}
\end{equation}
If we assume that only the mean term is learnable during training, and $\tilde{\sigma}_{\theta, t-1} = \tilde{\sigma}_{t-1}$, we can ignore all the unlearnable constants and arrive at the final training objective:
\begin{equation}
\begin{aligned}
    \mathcal{L} &= \mathbb{E}_{t, \mathbf{z}_0, \mathbf{z}_t, \mathbf{z}_T} \left[ \frac{1}{2 \tilde{\sigma}_t^2} \left\| \tilde{\boldsymbol{\mu}}_{t-1} - \tilde{\boldsymbol{\mu}}_{\theta, t-1}(\mathbf{z}_t, \mathbf{z}_T, t) \right\|^2 \right] \\
    &= \mathbb{E}_{t, \mathbf{z}_0, \mathbf{z}_t, \mathbf{z}_T} \left[ \left\| \boldsymbol{\epsilon} - \boldsymbol{\epsilon}_{\theta}(\mathbf{z}_t, \mathbf{z}_T, t) \right\|^2 \right]   
\end{aligned}
\end{equation}
which is the same as in \cref{eqn: diffusion loss}. This concludes our derivation.

\section{Training and Inference Pipeline}
\subsection{Pipeline Pseudo-code}
\begin{algorithm}[!h]
\caption{Training}
\label{alg:training}
\begin{algorithmic}[1]
\REQUIRE 
    Encoder $\mathcal{E}$; Decoder $\mathcal{D}$; Loss weights $\omega_1, \omega_2$
\REPEAT
    \STATE Retrieve paired data $(\mathbf{x}, \mathbf{y})$ 
    \STATE $\mathbf{z}_0=\mathcal{E}(\mathbf{y})$, $\mathbf{z}_T=\mathcal{E}(\mathbf{x})$
    \STATE $t \sim \text{Uniform}(\{1, \dots, T\})$, $\boldsymbol{\epsilon} \sim \mathcal{N}(\mathbf{0}, \mathbf{I})$
    \STATE $\mathbf{z}_t = m_t\mathbf{z}_0 + n_t\mathbf{z}_T + \sigma_t\boldsymbol{\epsilon}$
    \STATE $\mathbf{z}_t' = \frac{1}{\sqrt{n_t^2 + \sigma_t^2}}(\mathbf{z}_t - n_t \mathbf{z}_T)$
    \STATE $\mathcal{L}_{elbo} = \left\| \boldsymbol{\epsilon} - \boldsymbol{\epsilon}_{\theta}(\mathbf{z}_t', \mathbf{z}_T, t) \right\|^2,$
    \STATE $\hat{\mathbf{z}}_0 = \frac{1}{m_t}(\mathbf{z}_t - n_t \mathbf{z}_T - \sigma_t \boldsymbol{\epsilon}_{\theta}(\mathbf{z}_t', \mathbf{z}_T, t))$
    \STATE $\mathcal{L}_{ic} = \left\| \mathcal{D}(\hat{\mathbf{z}}_0) - \mathbf{y} \right\|^2$
    \STATE Take gradient step on $\nabla_\theta \left( \omega_1\mathcal{L}_{elbo} + \omega_2\mathcal{L}_{ic} \right)$
\UNTIL \textbf{Converged}
\end{algorithmic}
\end{algorithm}
\vspace{-0.5em}

\begin{algorithm}[!h]
\caption{Inference}
\label{alg:inference}
\begin{algorithmic}[1]
\REQUIRE 
    Encoder $\mathcal{E}$; Decoder $\mathcal{D}$; Trained $\boldsymbol{\epsilon}_{\theta}(\cdot, \cdot, \cdot)$
    
\STATE $\mathbf{z}_T=\mathcal{E}(\mathbf{x})$
\FOR{$t = T, \dots, 1$}
    \STATE $\boldsymbol{\epsilon} \sim \mathcal{N}(\mathbf{0}, \mathbf{I})$ if $t > 1$, else $\boldsymbol{\epsilon} = \mathbf{0}$
    \STATE $\mathbf{z}_t' = \frac{1}{\sqrt{n_t^2 + \sigma_t^2}}(\mathbf{z}_t - n_t \mathbf{z}_T)$
    \STATE $\hat{\mathbf{z}}_0 = \frac{1}{m_t}(\mathbf{z}_t - n_t \mathbf{z}_T - \sigma_t \boldsymbol{\epsilon}_{\theta}(\mathbf{z}_t', \mathbf{z}_T, t))$
    \STATE $\mathbf{z}_{t-1} = k_1 \mathbf{z}_t + k_2 \hat{\mathbf{z}}_0 + k_3 \mathbf{z}_T + \tilde{\sigma}_t \boldsymbol{\epsilon}$
\ENDFOR
\STATE $\mathbf{y} = \mathcal{D}(\hat{\mathbf{z}}_0)$ \\
\STATE \textbf{return} $\mathbf{y}$ 
\end{algorithmic}
\end{algorithm}

\subsection{Accelerate Inference}
Similar to DDIM~\cite{song2020denoising} and DBIM~\cite{zheng2024diffusion}, the inference process of DPBridge can be accelerated using non-Markovian sampling scheme that preserves the marginal distributions of intermediate states. Specifically, we perform sampling over a reduced set of timesteps $\{t_i\}_{i=1}^N$, where $0 = t_0 < t_1 < \cdots < t_{N-1} < t_N = T$, allowing the number of inference steps $N$ to be decoupled from the total number of training steps $T$. The accelerated inference process is given by:
\begin{equation}
\label{eqn: accelerate inference}
\begin{split}
    \mathbf{z}_{t_n} &= m_{t_n} \hat{\mathbf{z}}_0 + n_{t_n} \mathbf{z}_T + \\
    &\sqrt{\sigma_{t_n}^2 - g_n^2} \frac{\mathbf{z}_{t_{n+1}} - m_{t_{n+1}} \hat{\mathbf{z}}_0 - n_{t_{n+1}} \mathbf{z}_T}{\sigma_{t_{n+1}}} + g_n \boldsymbol{\epsilon}
\end{split}
\end{equation}
where $g_n$ denotes variance parameter that governs the stochasticity of the reverse sampling process. Notably, when $g_n=\tilde{\sigma}_t^2$, as defined \cref{eqn: bridge reverse kernel}, the process becomes a Markovian bridge.

\section{Network Architecture}
\label{app:network}
The basic structure of our network follows the original design of Stable Diffusion. To better utilize visual information ($\mathbf{z}_T$ in our case), we concatenating both intermediate samples and input image along the feature dimension and send in the noise prediction network as a whole. To address the dimension incompatibility with pretrained backbone, the input channels of the network are thus doubled to accommodate the concatenated input. In addition, as suggested by~\cite{ke2024repurposing}, we duplicate the weight tensor of these accommodated layers and divide the values by two to preserve the input magnitude to subsequent layers. As DPBridge is an image-conditioned generative framework, we employ null-text embedding for all cross-attention layers to eliminate the impact of text prompts.

\section{Implementation Details}
\label{app:implem details}
During training, the number of timesteps of is set to be 1000, and we use 50 sampling steps during inference with considerations of both quality and efficiency. Training our method takes 30K iterations using the Adam optimizer with a base learning rate of $3\times10^{-5}$ and an exponential decay strategy after 100-steps warm-up. The batch size is 2, but with 8 steps gradient accumulation, the effective batch size is equivalent to 16. We also apply random horizontal flipping augmentation during training. The whole process takes approximately 3 days to complete on a single Nvidia RTX 3090Ti GPU.

\section{Experiments}
\subsection{Additional Quantitative Results}
In this section, we further evaluate the robustness of DPBridge system by perturbing the input image with different types and different levels of intensity of noise. In this experiment setting, we use Stable Diffusion 1.5 as pretrained backbone. The evaluation is conducted under both depth estimation and surface normal prediction tasks. We use NYUv2 and KITTI to report the depth estimation results, and for surface normal prediction, we report the results on NYUv2 and iBims-1. \cref{tab:robustness eval} reveals that the model's performance degrades as noise intensity increases, but the model performance remains acceptable under moderate-level-intensity noise. Among noise types, uniform noise shows the mildest effect on degradation, while salt \& pepper and high-intensity Gaussian noise cause the most significant performance decline. In addition, the noise intensity has more impact on surface normal prediction than depth estimation, showcasing the inherent characteristics between different tasks. These results emphasize the model's robustness under moderate noise but still expose its vulnerability to extreme or specific types of noise.

\begin{table*}[h]
\centering
\resizebox{\linewidth}{!}{%
\begin{tabular}{lccccccccc}
\toprule
\multirow{3}{*}{ \textbf{Noise Type} } & \multirow{3}{*}{ \textbf{Noise Level} } & \multicolumn{4}{c}{\textbf{Depth Estimation}} & \multicolumn{4}{c}{\textbf{Surface Normal Prediction}}\\
&  & \multicolumn{2}{c}{\textbf{NYUv2}} & \multicolumn{2}{c}{\textbf{KITTI}} & \multicolumn{2}{c}{\textbf{NYUv2}} & \multicolumn{2}{c}{\textbf{iBims-1}}\\ 
&  & AbsRel$\downarrow$ & $\delta_1\uparrow$ & AbsRel$\downarrow$ & $\delta_1\uparrow$  & Mean$\downarrow$ & $11.25^{\circ}\uparrow$ & Mean$\downarrow$ & $11.25^{\circ}\uparrow$ \\
\midrule
    \textbf{Original Version} & - & 6.9 & 95.7 & 11.2 & 87.3 & 18.1 & 54.2 & 24.3 & 51.2 \\
\midrule
    \multirow{4}{*}{ \textbf{Gaussian} } 
    & $\mathcal{N}(0, 0.05)$ & 7.2 & 94.4 & 11.8 & 86.0 & 18.6 & 52.2 & 25.6 & 45.1 \\
    & $\mathcal{N}(0, 0.1)$ & 8.4 & 92.7 & 12.1 & 85.3 & 20.3 & 49.4 & 27.4 & 40.4 \\
    & $\mathcal{N}(0, 0.2)$ & 10.3 & 90.0 & 15.7 & 78.6 & 24.6 & 43.7 & 37.0 & 25.4 \\
    & $\mathcal{N}(0, 0.5)$ & 15.9 & 77.2 & 22.5 & 67.1 & 41.3 & 22.5 & 50.0 & 13.2 \\
\midrule
    \multirow{4}{*}{ \textbf{Uniform} }
    & $\mathcal{U}(-0.05, 0.05)$ & 7.0 & 95.0 & 11.4 & 86.5 & 18.3 & 53.6 & 24.5 & 50.7 \\
    & $\mathcal{U}(-0.1, 0.1)$ & 7.2 & 94.7 & 11.5 & 85.4 & 18.7 & 53.0 & 24.8 & 49.4 \\
    & $\mathcal{U}(-0.2, 0.2)$ & 7.6 & 94.0 & 12.0 & 85.0 & 19.5 & 50.5 & 26.0 & 46.1 \\
    & $\mathcal{U}(-0.5, 0.5)$ & 9.3 & 91.1 & 16.3 & 76.1 & 22.4 & 42.2 & 30.2 & 31.6 \\
\midrule
    \multirow{2}{*}{ \textbf{Poisson} }
    & \textit{Poisson}($\lambda=0.05$) & 10.2 & 90.2 & 13.1 & 82.5 & 21.3 & 45.6 & 28.1 & 38.9 \\
    & \textit{Poisson}($\lambda=0.1$) & 11.8 & 88.9 & 15.4 & 79.3 & 23.2 & 40.3 & 34.9 & 24.8\\
\midrule
    \multirow{2}{*}{ \textbf{Salt \& Pepper} }
    & \textit{Probability}=5\% & 12.0 & 85.4 & 13.3 & 81.7 & 26.3 & 28.8 & 31.8 & 29.6 \\
    & \textit{Probability}=10\% & 14.3 & 82.2 & 16.0 & 76.2 & 28.6 & 26.3 & 35.3 & 21.4\\
\bottomrule
\end{tabular}
}
\caption{Robustness evaluation on depth estimation and surface normal prediction tasks with different types and different intensities of noise. During all experiments, the input images are normalized between -1 to 1.}
\label{tab:robustness eval}
\end{table*}

\subsection{Additional Qualitative Results}
\begin{figure*}[!htbp]
    \begin{center}
    \includegraphics[width=0.82\linewidth]{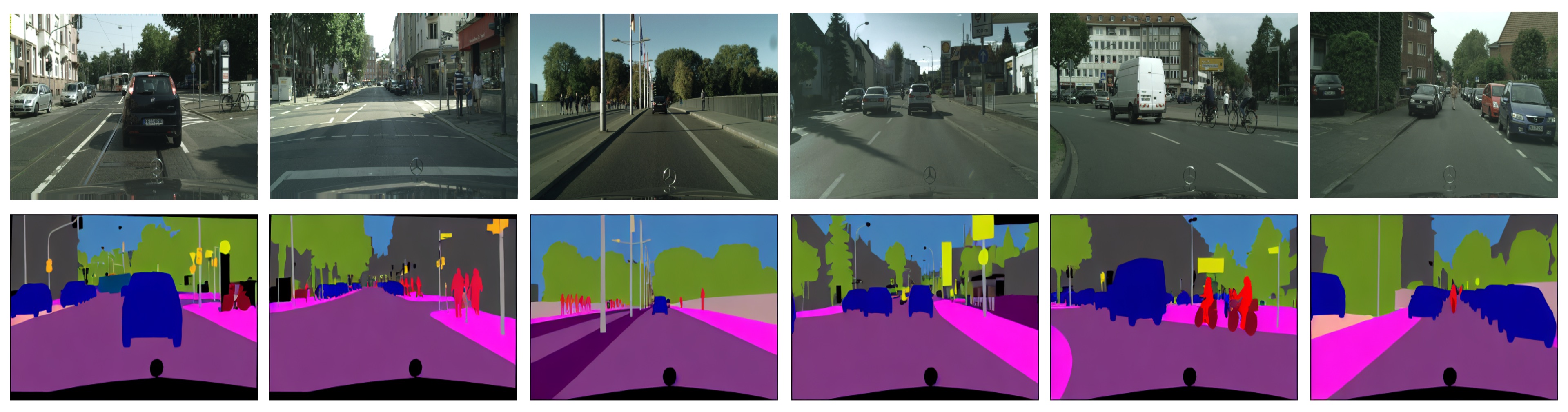}
    \caption{Qualitative examples of applying DPBridge to semantic segmentation task on CityScape~\cite{cordts2016cityscapes} dataset. We transform discrete labels into color maps and us our bridge model to predict colors in RGB space instead of categorical labels.}
    \label{fig:segm_cityscape}
    \end{center}
\vspace{-1.2em}
\end{figure*}

\begin{figure*}[!htbp]
    \begin{center}
    \includegraphics[width=0.82\linewidth]{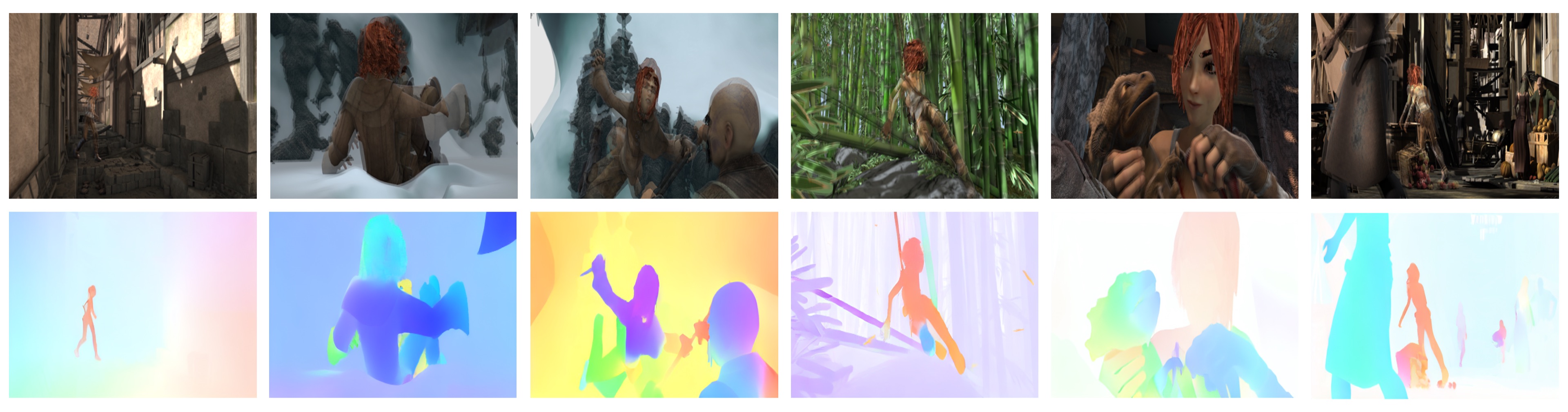}
    \caption{Qualitative examples of applying DPBridge to optical flow prediction task on MPI-Sintel~\cite{butler2012naturalistic} dataset. The input image is the overlayed RGB pair, and we transform the flow map into RGB space so that the diffusion bridge process can be constructed accordingly.}
    \label{fig:flow_sintel}
    \end{center}
\vspace{-1.2em}
\end{figure*}

\begin{figure*}[!htbp]
    \begin{center}
    \includegraphics[width=0.82\linewidth]{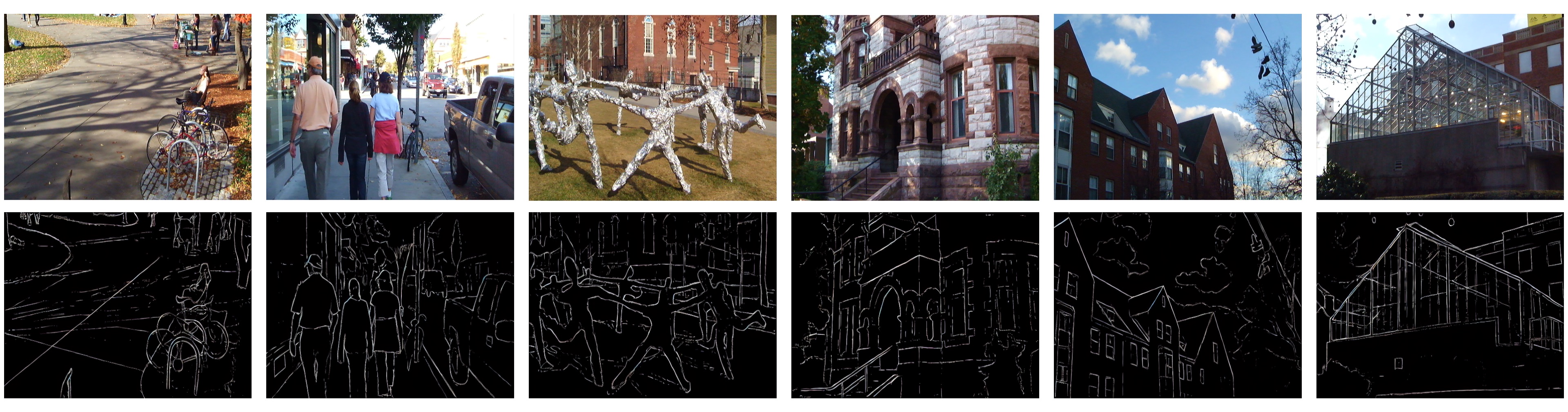}
    \caption{Qualitative examples of applying DPBridge to edge detection task on MultiCue~\cite{mely2016systematic} dataset. }
    \label{fig:edge_city}
    \end{center}
\vspace{-1.2em}
\end{figure*}

\begin{figure*}[!htbp]
    \begin{center}
    \includegraphics[width=\linewidth]{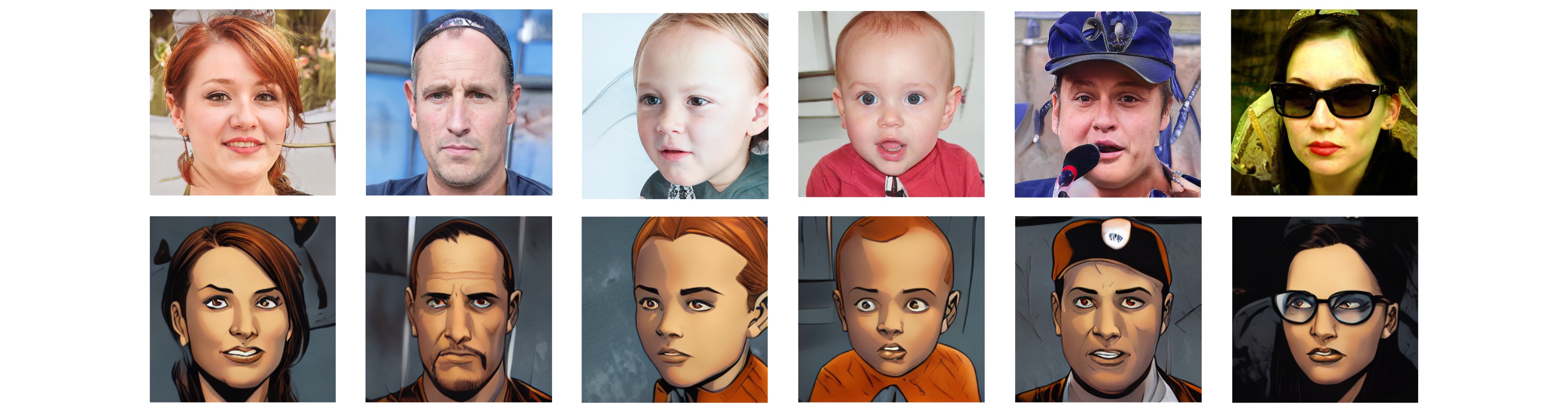}
    \caption{Qualitative examples of applying DPBridge to style transfer task on Face-to-Comics~\cite{face2comics} dataset.}
    \label{fig:face2comics}
    \end{center}
\vspace{-1.2em}
\end{figure*}

To further evaluate the effectiveness and generalization capability of DPBridge, we extend the framework to additional dense prediction tasks, including semantic segmentation, optical flow estimation, edge detection, and style transfer. Qualitative results are shown in \cref{fig:segm_cityscape}, \cref{fig:edge_city}, \cref{fig:flow_sintel}, and \cref{fig:face2comics}. These extensions are implemented by framing each task as a continuous-valued image-to-image translation problem and replacing the paired ground truth data accordingly. For semantic segmentation, we convert discrete labels into color maps and train the bridge model to predict colors in RGB space rather than class indices. For optical flow, we define the bridge process between the flow image and the first frame of the input RGB pair. To accommodate this, we modify the UNet architecture to condition on both RGB frames during inference and represent flow as a color-encoded image, enabling prediction in RGB space rather than raw uv coordinates. Across all tasks, DPBridge produces visually compelling results, demonstrating its versatility as a unified framework that can be easily adapted to a range of dense prediction scenarios. We note that these results are qualitative; quantitative evaluation and task-specific architectural refinements are left for future work.

\newpage

\end{document}